%% file: 00main.tex
\documentclass[final]{cvpr}

\usepackage{times}
\usepackage{epsfig}
\usepackage{graphicx}
\usepackage{amsmath}
\usepackage{amssymb}
\usepackage[numbers,sort]{natbib}
\usepackage[utf8]{inputenc} % allow utf-8 input
\usepackage[T1]{fontenc}    % use 8-bit T1 fonts
\usepackage[colorlinks=true,linkcolor=red, citecolor=green]{hyperref}
\usepackage{url}            % simple URL typesetting
\usepackage{booktabs}       % professional-quality tables
\usepackage{amsfonts}       % blackboard math symbols
\usepackage{amssymb}
\usepackage{nicefrac}       % compact symbols for 1/2, etc.
\usepackage{microtype}      % microtypography
\usepackage{adjustbox}
\usepackage{multirow}
\usepackage[flushleft]{threeparttable}
\usepackage{makecell}
\usepackage{tabulary}
\usepackage{amsmath}
\usepackage{tabularx}
\usepackage{pifont}
\usepackage{kotex}
\usepackage{bbold}
\usepackage{wrapfig}

\usepackage{enumitem}
\usepackage{graphicx}
\usepackage{scalerel}
\usepackage{caption}
\usepackage{tabularx}
\usepackage[linesnumbered,ruled,vlined]{algorithm2e}
\usepackage{stfloats}

\def\cvprPaperID{4016} % *** Enter the CVPR Paper ID here
\def\confYear{CVPR 2021}

\begin{document}

%%%%%%%%% TITLE
\title{Anti-Adversarially Manipulated Attributions for Weakly and Semi-Supervised Semantic Segmentation}

\author{Jungbeom Lee$^1$ ~~~~~~~ Eunji Kim$^1$ ~~~~~~~  Sungroh Yoon$^{1, 2, }$\thanks{Correspondence to: Sungroh Yoon <sryoon@snu.ac.kr>.}\\
$^1$ Department of Electrical and Computer Engineering, Seoul National University, Seoul, South Korea\\
$^2$ ASRI, INMC, ISRC, and Institute of Engineering Research, Seoul National University\\
{\tt\small \{jbeom.lee93, kce407, sryoon\}@snu.ac.kr}}

% \and

\maketitle

%%%%%%%%% ABSTRACT
\begin{abstract}
\vspace{-0.3em}
Weakly supervised semantic segmentation produces a pixel-level localization from a classifier, but it is likely to restrict its focus to a small discriminative region of the target object. AdvCAM is an attribution map of an image that is manipulated to increase the classification score. This manipulation is realized in an anti-adversarial manner, which perturbs the images along pixel gradients in the opposite direction from those used in an adversarial attack. It forces regions initially considered not to be discriminative to become involved in subsequent classifications, and produces attribution maps that successively identify more regions of the target object. 
In addition, we introduce a new regularization procedure that inhibits the incorrect attribution of regions unrelated to the target object and limits the attributions of the regions that already have high scores.
% restricts high attributions on discriminative regions of the object.
% excessive concentration of attributions on a small region of that object. 
% Our method is a post-hoc analysis of a trained classifier, which does not need to be altered or retrained. 
On PASCAL VOC 2012 test images, we achieve mIoUs of 68.0 and 76.9 for weakly and semi-supervised semantic segmentation respectively, which represent a new state-of-the-art.
The code is available at: \url{https://github.com/jbeomlee93/AdvCAM}.

\end{abstract}

\vspace{-0.8em}
%%%%%%%%% BODY TEXT
\section{Introduction}\label{intro}
Semantic segmentation involves the allocation of a semantic label to each pixel of an image.
It is an essential task in image recognition and scene understanding.
Deep neural networks (DNNs) have facilitated tremendous progress in semantic segmentation~\cite{chen2017deeplab, huang2019ccnet}; but they require a large number of training images annotated with pixel-level labels. Preparing such a training dataset is very expensive: pixel-level annotation of images containing an average of 2.8 objects takes about 4 minutes~\cite{bearman2016s} per image, and a single large (2048$\times$1024) image depicting a complicated scene requires more than 90 minutes for pixel-level annotation~\cite{cordts2016cityscapes}.

The need for pixel-level annotation is addressed by weakly supervised learning, in which a segmentation network is trained on images with less comprehensive annotations that are cheaper to obtain than pixel-level labels. 
Weakly supervised methods can use scribbles~\cite{tang2018normalized}, points~\cite{bearman2016s}, bounding boxes~\cite{khoreva2017simple, song2019box}, and class labels~\cite{lee2019ficklenet, Shimoda_2019_ICCV, ahn2018learning, chang2020weakly} as annotations. 
Labeling an image with class labels takes about 20 seconds~\cite{bearman2016s}, making class labels the cheapest option. 
In addition, many public datasets are already annotated with class labels~\cite{deng2009imagenet, everingham2010pascal}, and automated web searches can also provide images with class labels~\cite{lee2019frame, hong2017weakly, shen2018bootstrapping} although the accuracy of such labels may be low.
These considerations make class labels the most popular form of weak supervision.

\input{latex/Figures/overview}
Most weakly supervised segmentation methods that use class labels depend on attribution maps obtained from a trained classifier~\cite{zhou2016learning, selvaraju2017grad}.
Such a map identifies the image regions on which the classifier concentrated.
However, these important, or discriminative, regions are relatively small, and most attribution maps do not represent the whole region occupied by a target object, which makes those attribution maps unsuitable for training a semantic segmentation network.
Therefore, many researchers have tried to extend regions to cover more of a target object, by manipulating images~\cite{wei2017object, li2018tell, singh2017hide} or feature maps~\cite{lee2019ficklenet, zhang2018adversarial, hou2018self}.

One popular method for manipulation is erasure: the classifier is forced to find new regions of the target object from which discriminative regions previously located have been removed.
Erasure is effective, but it requires modification of the network, often by adding additional layers~\cite{hou2018self, zhang2018adversarial}, or additional training steps~\cite{wei2017object}.
Another difficulty is the provision of a reliable termination condition for the iterative erasure; the erasure of discriminative region of an image can cause the DNN to misclassify that image.
If the image from which the discriminative region has been erased crosses the decision boundary as shown in Figure~\ref{overview}(a), an erroneous attribution map may be generated.
An alternative method for manipulation is a stochastic perturbation shown in Figure~\ref{overview}(b). 
FickleNet~\cite{lee2019ficklenet} diversifies attribution maps from an image by applying random dropout to the feature maps of a DNN and aggregates them into a unified map.

We propose a new manipulation method for extending the discriminative regions of a target object.
Our method is based on adversarial attack~\cite{goodfellow2014explaining, kurakin2016adversarial}, but with a benign purpose. Adversarial attack finds a small perturbation of an image that pushes it across the decision boundary to change the classification result.
By contrast, our method operates in an anti-adversarial manner , which is the reversal of adversarial attack.
It aims to find a perturbation that pushes the manipulated image away from the decision boundary, as shown in Figure~\ref{overview}(c).
This manipulation is realized by adversarial climbing, in which an image is perturbed along pixel gradients which increase the classification score of the target class.
The result is that non-discriminative regions, which are nevertheless relevant to that class, gradually become involved in the classification, so that the CAM of the manipulated image identifies more regions of the object. Figure~\ref{overview}(d) shows examples of CAMs obtained by applying this manipulation technique iteratively.

Ascending the gradient ensures that classification score increases, but the repetitive ascending may cause irrelevant areas, such as parts of the backgrounds or regions of other objects, to be activated together or the attribution scores of some part of the target object to be increased dramatically.
We can address these problems by introducing regularization terms that suppress the scores of other classes and limit the attribution scores of the regions that already have high scores. 
The attribution maps obtained from images that have been iteratively manipulated in this way can be used as pseudo ground-truth masks to train a semantic segmentation network in a weakly and semi-supervised manner.

Our method is a post-hoc analysis of the trained classifier, and can be used to improve the performance of existing methods without modification, resulting in new state-of-the-art performance on the PASCAL VOC 2012 benchmark in both weakly and semi-supervised semantic segmentation.

The main contributions of this paper are three-fold:
\begin{itemize}
\vspace{-5pt}
	\item[$\bullet$] We propose AdvCAM, an attribution map of an image that is manipulated to increase the classification score, allowing it to identify more regions of an object.
	\vspace{-5pt}
	\item[$\bullet$] We empirically demonstrate that our method improves the performance of several methods of weakly supervised semantic segmentation without modification or re-training of their networks.
	\vspace{-5pt}
	\item[$\bullet$] Our technique produces significantly better performance on the Pascal VOC 2012 benchmark than existing methods, in both weakly and semi-supervised semantic segmentation.
\vspace{-3pt}
\end{itemize}

\section{Related Work}
\subsection{Weakly Supervised Learning}\label{re_weak}
Existing weakly supervised semantic segmentation methods aim to find the whole region occupied by a target object by obtaining an improved initial seed which contains a good approximation of the region occupied by the object, and growing that region so that more of the object is identified.

\textbf{Obtaining a High Quality Seed:}
Several methods have been proposed to improve the quality of the initial seeds obtained from classifiers. 
Wang \textit{et al.}~\cite{wang2020self} use equivariance regularization during the training of their classifier so that the attribution maps obtained from differently transformed images are equivariant to those transformations.
Chang \textit{et al.}~\cite{chang2020weakly} improve feature learning by using latent semantic classes that are sub-categories of annotated parent classes, which can be pseudo-labeled by clustering image features.
Fan \textit{et al.}~\cite{fan2018cian} and Sun \textit{et al.}~\cite{sun2020mining} capture information shared between several images by considering cross-image semantic similarities and differences. 
Wei \textit{et al.}~\cite{wei2018revisiting} and Lee \textit{et al.}~\cite{lee2018robust} consider the target object in several contexts by combining multiple attribution maps from differently dilated convolutions or from different layers of a DNN.

\textbf{Growing the Object Region:}
Some researchers expand an initial CAM~\cite{zhou2016learning} seed using a method analogous to region growing by examining the neighborhood of each pixel.
Semantic labels are propagated from regions which can confidently be associated with the target object to regions which were initially ambiguous.
SEC~\cite{kolesnikov2016seed} and DSRG~\cite{huang2018weakly} start with a initial CAM seed containing ambiguous regions, and allocates pseudo labels to those ambiguous region during the training of the segmentation network. 
PSA~\cite{ahn2018learning} and IRN~\cite{ahn2019weakly} extend the object region to semantically similar areas by a random walk.
BEM~\cite{chenweakly} synthesizes a pseudo boundary from a CAM and then uses a similar propagation with PSA~\cite{ahn2018learning}.

\subsection{Semi-Supervised Learning}\label{re_semi}
In semi-supervised learning, a segmentation network is trained using a small number of images with pixel-level annotations, together with a much larger number of images with weak annotations or none at all.
Cross-consistency training (CCT)~\cite{ouali2020semi} involves the training of a segmentation network with unlabeled, or weakly labeled, images by enforcing an invariance of the predictions over different perturbations, such as injecting random noise.
Souly \textit{et al.}~\cite{souly2017semi} improve feature learning by using images synthesized by generative adversarial network~\cite{goodfellow2014generative}.
Hung \textit{et al.}~\cite{hung2019adversarial} adopt adversarial training scheme that reduces the distribution gap between predicted segmentation maps and ground-truth maps.

\subsection{Adversarial Attack}\label{adv_manipulation}
Methods of adversarial attack attempt to fool a DNN by presenting it with manipulated input with the intent to deceive.
Adversarial attack can be applied to classification~\cite{goodfellow2014explaining, moosavi2016deepfool}, semantic segmentation~\cite{arnab2018robustness}, and object detection~\cite{xie2017adversarial}.
Deceptive attribution maps can also be produced by adversarial image manipulation~\cite{dombrowski2019explanations} or model parameter manipulation~\cite{heo2019fooling}.
The aim of such attacks is to replace an attribution map with a spurious map, which highlights another location in the same image, without significantly changing the output of the DNN.
Those methods are interested in manipulating the image to cause the neural network's unintended behavior. 
By contrast, we are interested in finding the proper manipulation of the input image, so the resulting attribution map can cover the target object better.

\section{Proposed Method}
We look more closely at adversarial attack methods and class activation map in Section~\ref{adv_attack_method}.
In Sections~\ref{Advcam_method} and~\ref{reg_sec}, we introduce AdvCAM and explain how we generate pseudo ground truth for weakly supervised semantic segmentation. Finally, we show how to train a semantic segmentation network with generated pseudo ground-truth in Section~\ref{train_segnet}.

\subsection{Preliminaries}\label{adv_attack_method}
\noindent\textbf{Adversarial Attack in more detail:} An adversarial attack aims to find a small pixel-level perturbation that can change the output from a DNN.
In other words, given an input $x$, it finds the perturbation $n$ satisfying $\texttt{NN}(x) \neq \texttt{NN}(x+n)$, where $\texttt{NN}(\mathord{\cdot})$ is the output of the neural network.
A representative method~\cite{goodfellow2014explaining} of constructing $n$ is to consider the normal vector to the decision boundary of $\texttt{NN}$, which can be realized by finding the gradients of $\texttt{NN}$ with respect to $x$. 
A manipulated image $x'$ can then be obtained as follows:
\vspace{-0.2em}
\begin{align}\label{adv_attack}
x' = x - \xi \nabla_x \texttt{NN}(x),
\vspace{-0.8em}
\end{align}
where $\xi$ determines the extent of the change to the image. This process can be understood as gradient descent.

\noindent\textbf{Class Activation Map (CAM):}
It identifies the region of an image which a classifier has used.
A CAM is computed from the class-specific contribution of each channel of the feature map to the classification score. It is based on a convolutional neural network that has global average pooling (GAP) before the last classification layer. 
A class activation map $\texttt{CAM}(x)$ from an image $x$ can be computed as follows:
\vspace{-0.4em}
\begin{equation}\label{cam}
\texttt{CAM}(x) = \mathbf{w}^\intercal_c f(x),
\vspace{-0.4em}
\end{equation}
where $\mathbf{w}_c$ is the weights of the final classification layer for class $c$, and $f(x)$ is the feature map of $x$ prior to GAP.

A CAM bridges the gap between image-level and pixel-level annotations. However, the regions obtained by a CAM are usually much smaller than the full extent of the target object, since the small discriminative regions provide sufficient information for classification.

\subsection{AdvCAM}\label{Advcam_method}
\subsubsection{Adversarial Climbing}
AdvCAM is an attribution map obtained through adversarial climbing, which is an anti-adversarial technique that manipulates the image so as to increase the classification score of that image, with the result that the classifier identifies more regions of objects.
This is the reverse of an adversarial attack based on Eq.~\ref{adv_attack}, which manipulates the image to reduce the classification score. 
Inspired by PGD~\cite{kurakin2016adversarial}, iterative adversarial climbing of the initial image $x^{0}$ can be performed using the following relation:
\vspace{-0.1em}
\begin{equation}\label{sgd}
x^{t} = x^{t-1} + \xi \nabla_{x^{t-1}} y_c^{t-1},
\vspace{-0.2em}
\end{equation}
where $t$ ($1\leq t \leq T$) is the adversarial step index, $x^{t}$ is the manipulated image at the $t-$th step, and $y_c^{t-1}$ is the classification logit of $x^{t-1}$ for class $c$. 

This process makes the previously non-discriminative yet relevant features become more involved in the classification.
Thus, the CAMs obtained from successive images manipulated by the iteration can be expected to identify an increasing amount of the region of the target object.
We produce a localization map $\mathcal{A}$ which encapsulates the results of the iteration by aggregating the CAMs obtained from the manipulated images at each iteration $t$, as follows: 
\vspace{-0.3em}
\begin{align}\label{aggregate}
\mathcal{A} = 
\frac{\sum_{t=0}^{T} \texttt{CAM}(x^{t})}{\max \sum_{t=0}^{T} \texttt{CAM}(x^{t})}.
\vspace{-0.5em}
\end{align}

\subsubsection{How can Adversarial Climbing Improve CAMs?}\label{how_advcam}

The connection between a classification logit $y_c$ and a CAM, \textit{i.e.} $y_c = \text{GAP}(\texttt{CAM})$~\cite{zhang2018adversarial}, infers that adversarial climbing increases $y_c$, and thus the CAM. 
In this process, features involved in classification are enhanced.
To provide a better understanding how adversarial climbing generates a denser CAM, we consider two questions: \textcircled{\raisebox{-0.9pt}{1}} Can non-discriminative features be enhanced? \textcircled{\raisebox{-0.9pt}{2}} Are those enhanced features class-relevant from a human point of view?

\textbf{\textcircled{\raisebox{-0.9pt}{1}} Can non-discriminative features be enhanced?:} One might think that changing a pixel with a large gradient primarily enhances discriminative features. 
This pixel change affects many features due to the receptive field. However, not all the affected features are necessarily discriminative.
We support this analysis empirically.
We define the discriminative region $R_\text{D} \!=\! \{i|\texttt{CAM}(x^{0})_i\!\geq\!0.5\}$ and the non-discriminative region $R_{\text{ND}} \!=\! \{i|0.1\!<\!\texttt{CAM}(x^{0})_i\!<\!0.5\}$, where $i$ is the location index. 
The pixel amplification ratio $s^i_t$ is $\texttt{CAM}(x^{t})_i/\texttt{CAM}(x^{0})_i$ at location $i$ and step $t$.
Figure~\ref{fig_amp}(a) shows that adversarial climbing makes both $s^{i \in {R_\text{D}}}_t$ and $s^{i \in {R_{\text{ND}}}}_t$ grow, but enhances non-discriminative features more than discriminative ones, resulting in a denser CAM.

\input{latex/Figures/figure_amplification}

\textbf{\textcircled{\raisebox{-0.9pt}{2}} Are those enhanced features class-relevant from a human point of view?}
We now consider whether the highlighted non-discriminative features are class-relevant from a human point of view.
Moosavi \textit{et al.}~\cite{moosavi2019robustness} argued that a loss landscape that is sharply curved with respect to input makes a NN vulnerable to adversarial attack.
Researchers have subsequently shown that a flattened loss landscape, obtained by reducing the curvature of the loss surface~\cite{moosavi2019robustness} or encouraging the loss to behave linearly~\cite{qin2019adversarial}, can improve the robustness of a NN.
Systems which are robust in this sense have been shown to produce features that align better with human perception and operate in a easier way to understand~\cite{santurkar2019image, tsipras2018robustness, ilyas2019adversarial}.

By the same token, we can expect that images manipulated by adversarial climbing will produce features that align with human perception well because the curvature of loss surface affected by adversarial climbing is small.
To support this, we visualize the loss landscape of our trained classifier, following Moosavi \textit{et al.}~\cite{moosavi2019robustness}: we obtain a manipulation vector $\vec{n}$ and a random vector $\vec{r~}$ from the classification loss $\ell$ computed from an image. 
We determine the surfaces of classification loss values computed from images, manipulated by a vector which is interpolated between $\vec{n}$ and $\vec{r}~$ using a range of interpolation ratios.
The loss landscape obtained by adversarial climbing (Figure~\ref{landscape}(a)) is much more flatten than that obtained by adversarial attacking (Figure~\ref{landscape}(b)).
Therefore, we can legitimately expect it to increase the attribution of features relevant to the class from a human point of view, resulting in a better CAM.

\input{latex/Figures/Figure_landscapes}

\newcommand{\RNum}[1]{\lowercase\expandafter{\romannumeral #1\relax}}

\subsection{Regularization}\label{reg_sec}
Even if the loss surface obtained by adversarial climbing is reasonably flat, too much repetitive adversarial manipulation may cause regions corresponding to objects in the wrong class to be activated, or increase the attribution scores of the regions that already have high scores.
We address this by (\RNum{1}) suppressing the logit values associated with other classes and (\RNum{2}) restricting high attributions on discriminative regions of the target object.

\textbf{Suppressing Other Classes:}
In an image, objects of different classes can mutually increase logit values.
For example, since a chair and a dining table mainly occur together in an image, a NN may infer an increased logit value for the chair from the region of the table.
We thus add regularization that reduces logit values for all classes except $c$.

\textbf{Restricting High Attributions:}
As mentioned in Section~\ref{how_advcam}, adversarial climbing increases the attribution scores for both discriminative and non-discriminative regions in the feature map.
However, the growth of attribution scores for discriminative regions is problematic for two reasons: 1) it prevents new regions from being additionally attributed to the classification score, and 2) if the maximum value of the attribution score increases during adversarial climbing, the normalized scores of the remaining area may decrease. Please see the blue boxes in Figure~\ref{mask_ex}(b).

Therefore we limit the attribution scores in regions that already have high scores during adversarial climbing, so the attribution scores of those regions remain similar to that of $x^{0}$.
We realize this scheme by introducing a restricting mask $\mathcal{M}$ that contains the regions whose attribution scores of $\texttt{CAM}(x^{t-1})$ are higher than the threshold $\tau$. 
More specifically, $\mathcal{M}$ can be represented as follows:
\vspace{-0.1em}
\begin{equation}\label{mask}
\mathcal{M} = \mathbb{1}(\texttt{CAM}(x^{t-1}) > \tau),
\vspace{-0.1em}
\end{equation}
where $\mathbb{1}(\mathord{\cdot})$ is an indicator function. An example mask $\mathcal{M}$ is shown in Figure~\ref{mask_ex}(a). 

We add the regularization term so that the values of the CAM corresponding to the regions of $\mathcal{M}$ are forced to equal to that of $\texttt{CAM}(x^{0})$. 
With this regularization, $s^{i \in {R_\text{D}}}_t$ remains fairly constant but $s^{i \in {R_\text{ND}}}_t$ still grows during adversarial climbing (Figure~\ref{fig_amp}(b)).
Figure~\ref{fig_amp} shows that, adversarial climbing enhances non-discriminative features more than discriminative features (< 2$\times$), and regularization makes this difference even
larger (> 2.5$\times$).
Thus, new regions of the target object are found more effectively, resulting in a denser CAM (Figure~\ref{mask_ex}(b)).

To apply regularization, we modify Eq.~\ref{sgd} as follows:
\vspace{-0.3em}
\begin{flalign}\label{sgd_reg}
~~~~~~~~~~~~&x^{t} = x^{t-1} + \xi \nabla_{x^{t-1}} \mathcal{L}, ~~\text{where}&
\end{flalign}
\vspace{-2.2em}
\begin{align}\label{reg_loss}
\begin{split}
\mathcal{L} = &~ y_c^{t-1} - \sum_{k \in \mathcal{C} \setminus c}  y_k^{t-1} \\\\[-2em] &-  \lambda \left\lVert \mathcal{M} \odot |\texttt{CAM}(x^{t-1}) -  \texttt{CAM}(x^{0})|\right\lVert_1.
\vspace{-0.5em}
\end{split}
\end{align}
$\mathcal{C}$ is the set of all classes, $\lambda$ is a hyper-parameter that controls the influence of masking regularization, and $\odot$ is element-wise multiplication.

\subsection{Training Segmentation Networks}\label{train_segnet}
Since CAM is obtained from down-sampled intermediate features produced by the classifier, it localizes the target object coarsely and cannot represent its exact boundary.
Many methods of generating an initial seed for weakly supervised semantic segmentation construct a pseudo ground-truth by modifying their initial seeds using existing seed refinement methods~\cite{huang2018weakly, ahn2018learning, ahn2019weakly}. For example, SEAM~\cite{wang2020self} and Chang \textit{et al.}~\cite{chang2020weakly} use PSA~\cite{ahn2018learning}; and MBMNet~\cite{liu2020weakly} and CONTA~\cite{zhang2020causal} use IRN~\cite{ahn2019weakly}. 
We also apply the seed refinement method to the coarse map $\mathcal{A}$. For weakly supervised learning, we use the resulting profiles as pseudo ground-truth for training DeepLab-v2, pre-trained on the ImageNet dataset~\cite{deng2009imagenet}. 
For semi-supervised learning, we employ CCT~\cite{ouali2020semi}, which uses IRN~\cite{ahn2019weakly} to generate pseudo-ground truth masks; we replace these with our masks, constructed as just described.

\input{latex/Figures/figure_masking}

\vspace{-0.3em}
\section{Experiments}
\vspace{-0.2em}
\subsection{Experimental Setup}\label{setup_sec}
\vspace{-0.2em}
\textbf{Dataset:} We conducted experiments on the PASCAL VOC 2012~\cite{everingham2010pascal} dataset.
The images in this dataset come with masks for fully supervised semantic segmentation, but we only used them for evaluation. 
In a weakly supervised setting, we trained our network on 10,582 training images provided by Hariharan \textit{et al.}~\cite{hariharan2011semantic}, which have image-level annotations. 
In a semi-supervised setting, we used 1,464 training images with pixel-level annotations and 9,118 training images with class labels, following previous works~\cite{lee2019ficklenet, ouali2020semi, wei2018revisiting, luosemi}.
We evaluated our results by calculating mean intersection-over-union (mIoU) values for 1,449 validation images and 1,456 test images. Since the labels for test images are not publicly available, the results for those images were obtained from the official PASCAL VOC evaluation server. 

\textbf{Reproducibility:}
We performed iterative adversarial climbing with $T=27$ and $\xi=0.008$. We set $\lambda$ to 7 and $\tau$ to 0.5.
To generate the initial seed, we followed the procedure of Ahn \textit{et al.}~\cite{ahn2019weakly}, including the use of ResNet-50~\cite{he2016deep}.
For final segmentation, we used DeepLab-v2-ResNet101~\cite{chen2017deeplab} as the backbone network. We followed the default settings of~\cite{chen2017deeplab} for training, which included cropping the images to 321$\times$321 pixels.
In a semi-supervised setting we used the same settings as Ouali \textit{et al.}~\cite{ouali2020semi}.

\vspace{-0.2em}
\subsection{Experimental Results}
\vspace{-0.2em}
{\textbf{Quality of the Mask:}} 
Table~\ref{table_seed} compares the initial seed and pseudo ground truth masks obtained by our method and by other recent techniques. Both seeds and masks were generated from training images of the PASCAL VOC dataset. 
For initial seeds, we report the best results by applying a range of thresholds to separate the foreground and background in the map $\mathcal{A}$, as following SEAM~\cite{wang2020self}.
Our initial seeds are 6.8\% better than the original CAMs~\cite{zhou2016learning}, which provide a baseline, and this also outperforms the other methods.
Note that Chang \textit{et al.}~\cite{chang2020weakly} and SEAM~\cite{wang2020self} use Wide ResNet-38~\cite{wu2019wider}, which provides better representation than ResNet-50~\cite{he2016deep}. SEAM~\cite{wang2020self} also uses an auxiliary self-attention module that performs pixel-level refinement of the initial CAM by considering the relationship between pixels. 
We apply CRF, a widely used post-processing method, to the initial seeds of Chang \textit{et al.}~\cite{chang2020weakly}, SEAM~\cite{wang2020self}, IRN~\cite{ahn2019weakly}, and our method.
With the exception of SEAM, CRF improves the seed by more than 5\% on average, but it improves the seed of SEAM only by 1.4\%.
We believe this is because the seed of SEAM is already refined by the self-attention module. Our seed after applying CRF is 5.3\% better than that of SEAM.

We also compared pseudo ground truth masks, extracted after seed refinement, with existing methods.
Most methods refine their initial seeds with PSA~\cite{ahn2018learning} or IRN~\cite{ahn2019weakly}.
For a fair comparison, we produced pseudo ground truth masks using both these seed refinement techniques. 
Table~\ref{table_seed} shows that our method outperforms the others by a large margin, whichever seed refinement technique is used.

\input{latex/tables/table_seed}

{\textbf{Weakly Supervised Semantic Segmentation:}} Table~\ref{table_semantic} compares our method with other recently introduced weakly supervised semantic segmentation methods with various levels of supervision: fully supervised pixel-level masks ($\mathcal{P}$), bounding boxes ($\mathcal{B}$) or image class labels ($\mathcal{I}$), with and without salient object masks ($\mathcal{S}$). All the results in Table~\ref{table_semantic} were obtained using a ResNet-based backbone~\cite{he2016deep}.
With image-level annotation alone, our method achieves mIoU values of 68.1 and 68.0 for the PASCAL VOC 2012 validation and test images respectively. This is significantly better than the other methods under the same level of supervision. In particular, the mIoU value for validation images is 4.6\% higher than that for IRN~\cite{ahn2019weakly}, which is our baseline.
CONTA~\cite{zhang2020causal}, the best-performing method among our competitors, achieves an mIoU value of 66.1; but their method depends upon SEAM~\cite{wang2020self}, which is known to outperform IRN~\cite{ahn2019weakly}. If CONTA is implemented with IRN, the resulting mIoU value is 65.3, which is 2.8\% worse than our method. Figure~\ref{segsample} presents examples of semantic masks produced by FickleNet~\cite{lee2019ficklenet}, IRN~\cite{ahn2019weakly}, and our method.

Our method also outperforms other methods using auxiliary salient object mask supervision~\cite{li2014secrets, liu2010learning} that provides exact boundary information of salient objects in an image, or extra web images or videos~\cite{sun2020mining, lee2019frame}.
The performance of our method is also comparable with that of methods~\cite{song2019box, khoreva2017simple} that use bounding box supervision.

{\textbf{Semi-Supervised Semantic Segmentation:}} 
Table~\ref{tabsemi} compares the mIoU scores of our method on the PASCAL VOC validation and test images with those of other recent semi-supervised segmentation methods, which use 1.5K images with fully supervised masks and 9.1K images with weak annotations.
All the methods in Table~\ref{tabsemi} were implemented on the ResNet-based backbone~\cite{he2016deep}, except that daggered ($\dagger$) methods which used the VGG-based backbone~\cite{simonyan2014very}.
We achieve mIoU values of 77.8 and 76.9 for the PASCAL VOC 2012 validation and test images respectively, which is better than the other methods under the same level of supervision. 
Specifically, the performance of our method on the validation images was 4.6\% better than that of CCT~\cite{ouali2020semi}, which is our baseline. Our method even outperforms Song \textit{et al.}~\cite{song2019box} which uses bounding box labels for 9.1K images, instead of class labels.
Figure~\ref{segsample} presents examples of semantic masks produced by CCT~\cite{ouali2020semi} and our method.

\input{latex/tables/table_weak_sem_pascal}
\input{latex/tables/table_semi}

\input{latex/Figures/figure_seg_samples}
\input{latex/Figures/figure_each_iter}

\section{Discussion}

\subsection{Iterative Adversarial Climbing}\label{iterative}
We analyzed the effectiveness of the iterative adversarial climbing and regularization technique introduced in Section~\ref{reg_sec} by evaluating the initial seed in terms of mIoU. Figure~\ref{eachiter}(a) shows the mIoU of the initial seed for each adversarial iteration. Initially, the mIoU rises steeply, with or without regularization; but without regularization the curves peaks around iteration 8.

To analyze this, we evaluate the truthfulness of the newly localized region at each adversarial climbing iteration in terms of the proportion of noise, which we define to be the proportion of pixels that are classified as foreground but are actually background. Without regularization, the proportion of noise rises steeply after some iterations as shown in Figure~\ref{eachiter}(b), which means that new regions tend to be in the regions of background.
Regularization allows new regions of the target object to be found in as many as 30 adversarial steps, keeping the proportion of noise much lower than that of initial CAM.
Figure~\ref{eachiter_ex} shows examples of attribution maps at each adversarial iteration with and without regularization.

\input{latex/Figures/figure_each_iter_examples}

\input{latex/tables/table_pseudo_gt}

\subsection{Hyper-Parameter Analysis}\label{hyperparam}
In the previous section, we looked at the effect of the number of adversarial iterations (Figures~\ref{eachiter}(a) and (b)).
We also analyzed the sensitivity of the mIoU of the initial seed to the other three hyper-parameters used by AdvCAM.

\textbf{Regularization Coefficient $\boldsymbol{\lambda}$:} It controls the influence of the masking technique that limits the attribution scores of the regions that already have high scores during adversarial climbing, in Eq.~\ref{reg_loss}.
Figure~\ref{eachiter}(c) shows the mIoU of the initial seed for different values of $\lambda$. When $\lambda=0$, there is no regularization.
Masking technique improves performance by more than 5\% (50.43 for $\lambda=0$ \textit{vs.} 55.55 for $\lambda=7$).
The flattening of the curve after $\lambda=5$ suggests that it is not difficult to select a good value of $\lambda$.

\textbf{Masking Threshold $\boldsymbol{\tau}$}: It controls the size of the restricting mask $\mathcal{M}$ in Eq.~\ref{mask}, determining how many pixels' attribution values will remain similar to that of the original CAM during adversarial climbing.
Figure~\ref{eachiter}(d) shows the mIoU of the initial seed for different values of $\tau$. This parameter is even less sensitive than $\lambda$: varying $\tau$ between 0.3 and 0.7 produces less than 1\% change in mIoU.

\textbf{Step Size $\boldsymbol{\xi}$}: It determines the extent of the manipulation to the image in Eq.~\ref{sgd_reg}. Figure~\ref{eachiter}(e) shows the mIoU of the initial seed for different values of $\xi$. In our system, changes in step size $\xi$ are not particularly significant.

\subsection{Generality of Our Method}\label{generality}
In addition to IRN~\cite{ahn2019weakly}, we experimented with two state-of-the-art methods of generating an initial seed for weakly supervised semantic segmentation, namely Chang \textit{et al.}~\cite{chang2020weakly} and SEAM~\cite{wang2020self}. 
We used the authors' pre-trained classifier where possible, but we re-trained the classifier of IRN~\cite{ahn2019weakly} since the authors do not provide pre-trained one. We also followed their experimental settings including the backbone networks and mask refinement methods, \textit{i.e.,} we used PSA~\cite{ahn2018learning} to refine the initial seed from ``Chang \textit{et al.} + AdvCAM" or ``SEAM + AdvCAM".
Table~\ref{tab:baselines} gives mIoU values for the initial seed and the pseudo ground truth mask obtained by combining each method with adversarial climbing. The use of AdvCAM improves the quality of the initial seed by an average of over 4\%. 
Our approach does not require those initial seed generators to be modified or retrained.

\input{latex/Figures/figure_tsne}

\subsection{Manifold Visualization}
For visualizing a trajectory of adversarial climbing at a feature-level, we used t-SNE dimensional reduction~\cite{maaten2008visualizing}. 
We collect images that contain a single class of a cat or a bird and that are predicted by the classifier correctly. 
We then construct a set $\mathcal{F}$ containing the features of those images, before the final classification layer. 
We also choose a representative image of a cat, and another of a bird, and construct a set $\mathcal{F}~'$ containing the features of those two images and their 20 manipulated images by adversarial climbing.
Figure~\ref{fig_tsne} presents t-SNE visualization of features in $\mathcal{F} \cup \mathcal{F}~'$.
We can see that adversarial climbing actually pushes the features away from the decision boundary boundary that separates the blue and green areas.
In addition, despite 20 adversarial climbing steps, the manipulated features did not deviate significantly from the feature manifold of each class.

\section{Conclusion}
We have shown how adversarial manipulation can be used to expand the small discriminative regions of a target object, so as to obtain a better localization of that object.
We manipulate images with a pixel-level perturbation, which is obtained from the gradient computed from the output of classifier with respect to the input image, which increase the classification score of the perturbed image. The attribution map of the manipulated image covers more of the target object.
This is a post-hoc analysis of a trained classifier, and therefore no modification or re-training of the classifier is required.
This allows AdvCAM to be readily integrated into existing methods. We have shown that AdvCAM can indeed be combined with recent weakly supervised semantic segmentation networks, and achieved new state-of-the-art performance on both weakly and semi-supervised semantic segmentation. 

\bigskip
\noindent\textbf{Acknowledgements:}
This work was supported by the National Research Foundation of Korea (NRF) grant funded by the Korea government (MSIT) [2018R1A2B3001628], AIR Lab (AI Research Lab) in Hyundai \& Kia Motor Company through HKMC-SNU AI Consortium Fund, and the Brain Korea 21 Plus Project in 2021.

{\small
% \balance
\bibliographystyle{ieee_fullname}
\bibliography{egbib}
}

\setcounter{section}{0}
\renewcommand\thesection{\Alph{section}}
\setcounter{table}{0}
\renewcommand{\thetable}{A\arabic{table}}
\setcounter{figure}{0}
\renewcommand{\thefigure}{A\arabic{figure}}

\clearpage

\input{latex/Figures/fig_threshold}

\input{latex/tables/table_per_class}

\section{Appendix}
\subsection{Implementation Details}
\textbf{Details for Adversarial Climbing:}
Many recent studies~\cite{wang2020self, chang2020weakly, zhang2020causal} rely on the procedure of PSA~\cite{ahn2018learning} and IRN~\cite{ahn2019weakly} for generating a CAM: a single image is flipped and resized with four different scales of \{0.5, 1.0, 1.5, 2.0\}, and the CAMs are extracted from those eight images. Those CAMs are aggregated into a single map by pixel-wise sum pooling.
We manipulate those eight images independently for adversarial climbing.

\textbf{Details for Semantic Segmentation:} 
We used the PyTorch implementation of DeepLab-v2-ResNet101\footnote{\url{https://github.com/kazuto1011/deeplab-pytorch}} to train our segmentation network.
We used multi-scale testing during inference time following~\cite{wang2020self, ahn2019weakly, ahn2018learning, lee2019ficklenet, lee2019frame}. Specifically, an input image is resized with four different scales of \{0.5, 0.75, 1.0, 1.25\}. These images are fed into the segmentation network independently, and the outputs are aggregated into a single map by pixel-wise max pooling, resulting in the final segmentation map. The experiments were performed on NVIDIA Tesla V100 GPUs.

\subsection{Additional Analysis}

\textbf{Threshold analysis:} 
As mentioned in Section \textcolor{red}{4.2} of the main paper, we report the best initial seed performance by applying a range of thresholds to separate the foreground and background in the map $\mathcal{A}$. 
We present the effectiveness of this threshold by evaluating the initial seed, separated by a range of thresholds, in terms of mIoU. Figure~\ref{appendix_eachiter_ex}(a) shows the mIoU of the initial seed obtained from the `CAM', `AdvCAM without regularization', and `AdvCAM with regularization'. We select $t=8$ for `AdvCAM without regularization' and $t=27$ for `AdvCAM with regularization', which are the best values of $t$ for each setting according to Figure \textcolor{red}{6}(a) in the main paper.

\textbf{Effects of suppressing other classes:} Section \textcolor{red}{3.4} in the main paper has proposed two regularization terms: 1) suppressing other classes and 2) inhibiting excessive concentration.
The effectiveness of the latter was dealt with in-depth in the main paper (please see Section \textcolor{red}{5}). We will now focus on the effectiveness of suppressing other classes.
To isolate the effect of this regularization procedure, we exclude the masking technique in all experiments here.

Figure~\ref{appendix_eachiter_ex}(b) shows the mIoU of the initial seed for each adversarial iteration with and without the regularization of suppressing other classes. We can see that using this regularization technique provides better adversarial manipulation.

\textbf{Comparison of per-class mIoU scores:}
Table~\ref{class-specific-results} shows the per-class mIoU of our method and recently produced methods.

\textbf{Additional mask examples on semantic segmentation.}
Figure~\ref{appendix_segsample} shows more examples of the semantic masks from FickleNet~\cite{lee2019ficklenet}, IRN~\cite{ahn2019weakly}, CCT~\cite{ouali2020semi}, and our method.

\textbf{Additional examples of localization maps by adversarial climbing.}
Figure~\ref{appendix_eachiter_ex_successive} shows additional examples of successive attribution maps obtained from images manipulated by iterative adversarial climbing.

\input{latex/Figures/seg_ex}
\input{latex/Figures/seed_ex}

\end{document}

%% file: latex/Figures/overview.tex
\begin{figure}[t]
\centering
\includegraphics[width=\linewidth]{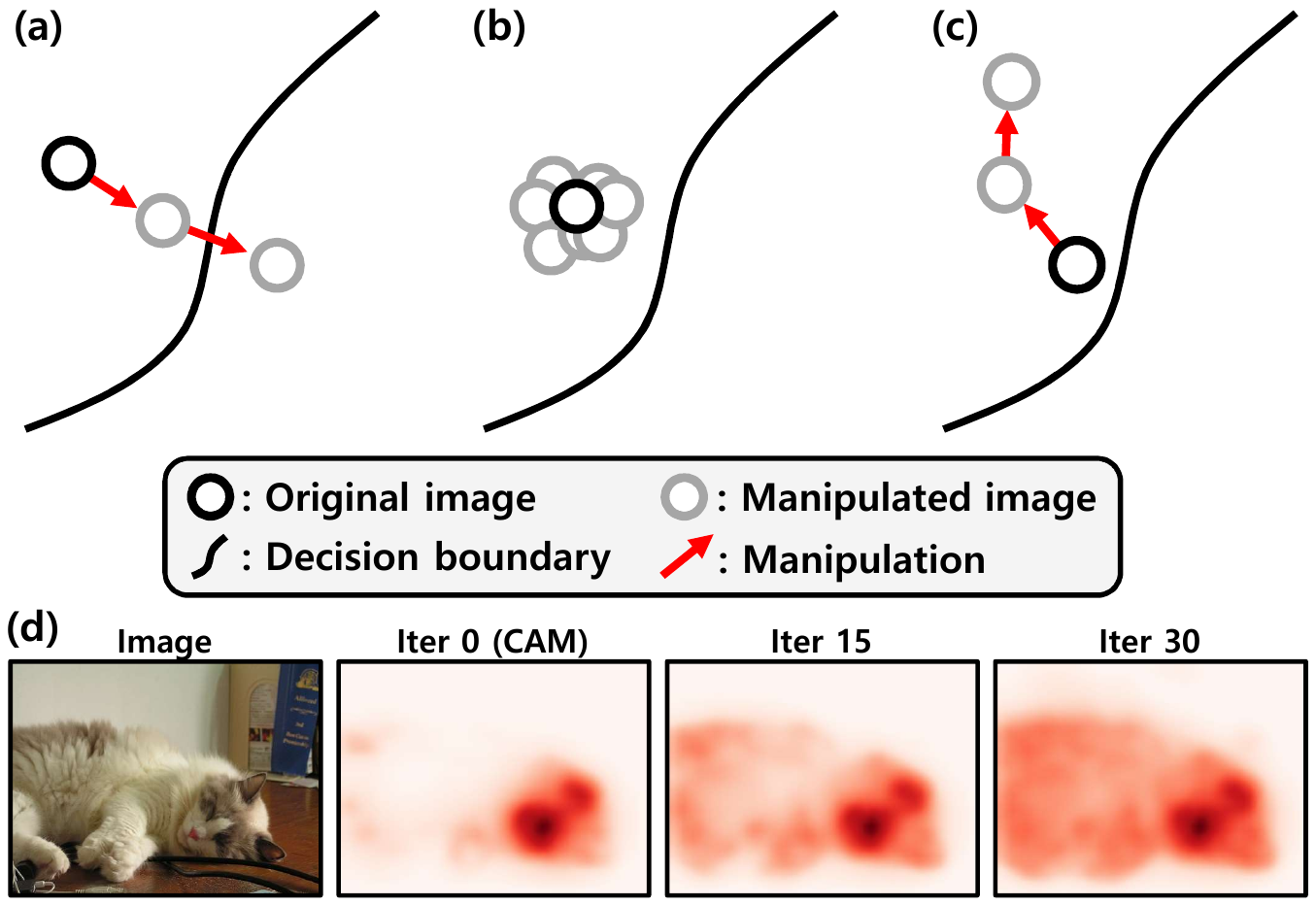}
\vspace{-1.5em}
\caption{\label{overview}  Conceptual description of image manipulation methods for weakly supervised semantic segmentation: (a) erasure~\cite{wei2017object, hou2018self, zhang2018adversarial}; (b) FickleNet~\cite{lee2019ficklenet}; and (c) AdvCAM. (d) Examples of successive attribution maps obtained from iteratively manipulated images.}
\vspace{-1.3em}
\end{figure}

%% file: latex/Figures/figure_amplification.tex
\begin{figure}[t]
\centering
\includegraphics[width=\linewidth]{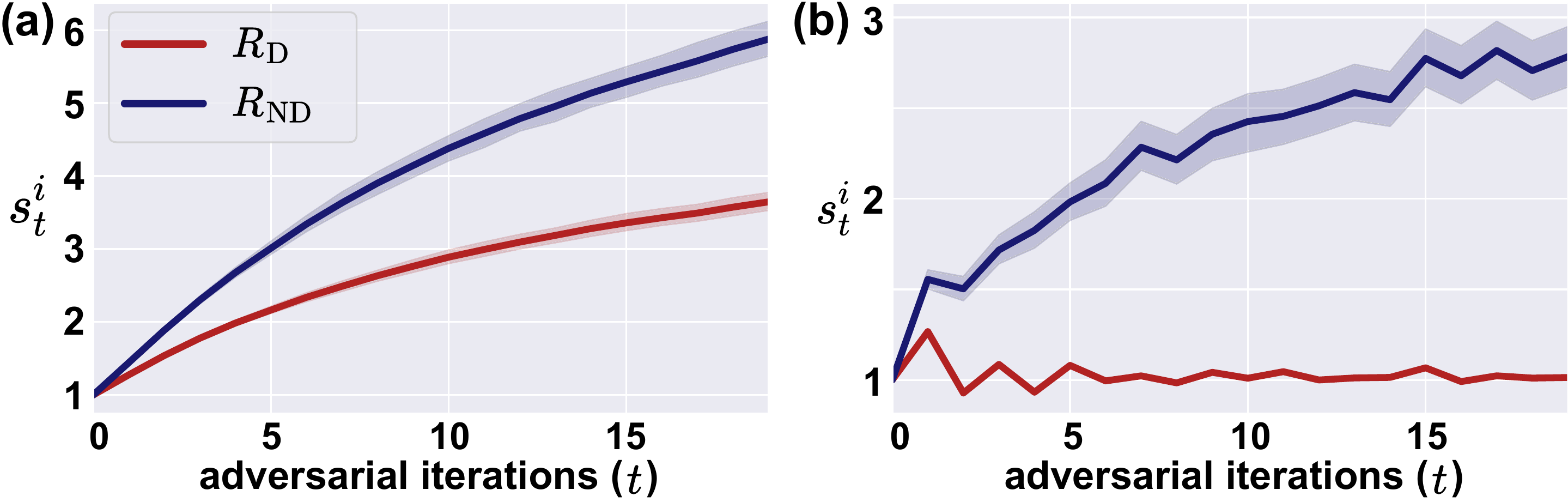}
\vspace{-2em}
\caption{\label{fig_amp} Distributions of the pixel amplification ratio $s^i_t$ for $i \in R_\text{D}$ and $i \in R_{\text{ND}}$ for 100 images, (a) without regularization and (b) with regularization.}
\vspace{-1em}
\end{figure}

%% file: latex/Figures/Figure_landscapes.tex
\begin{figure}[t]
\centering
\includegraphics[width=0.9\linewidth]{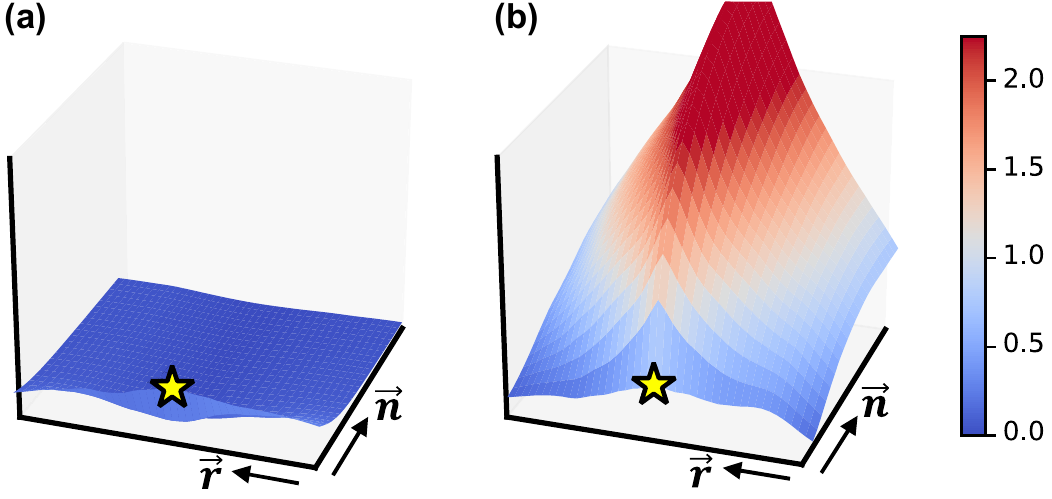}
\vspace{-.7em}
\caption{\label{landscape} Loss landscapes by manipulating images with weighted sums of the normal vector $\vec{n}~$ and a random vector $\vec{r}~$ for (a) adversarial climbing and (b) adversarial attack. The yellow star corresponds to the original image.}
\vspace{-.6em}
\end{figure}

%% file: latex/Figures/figure_masking.tex
\begin{figure}[t]
\centering
\includegraphics[width=\linewidth]{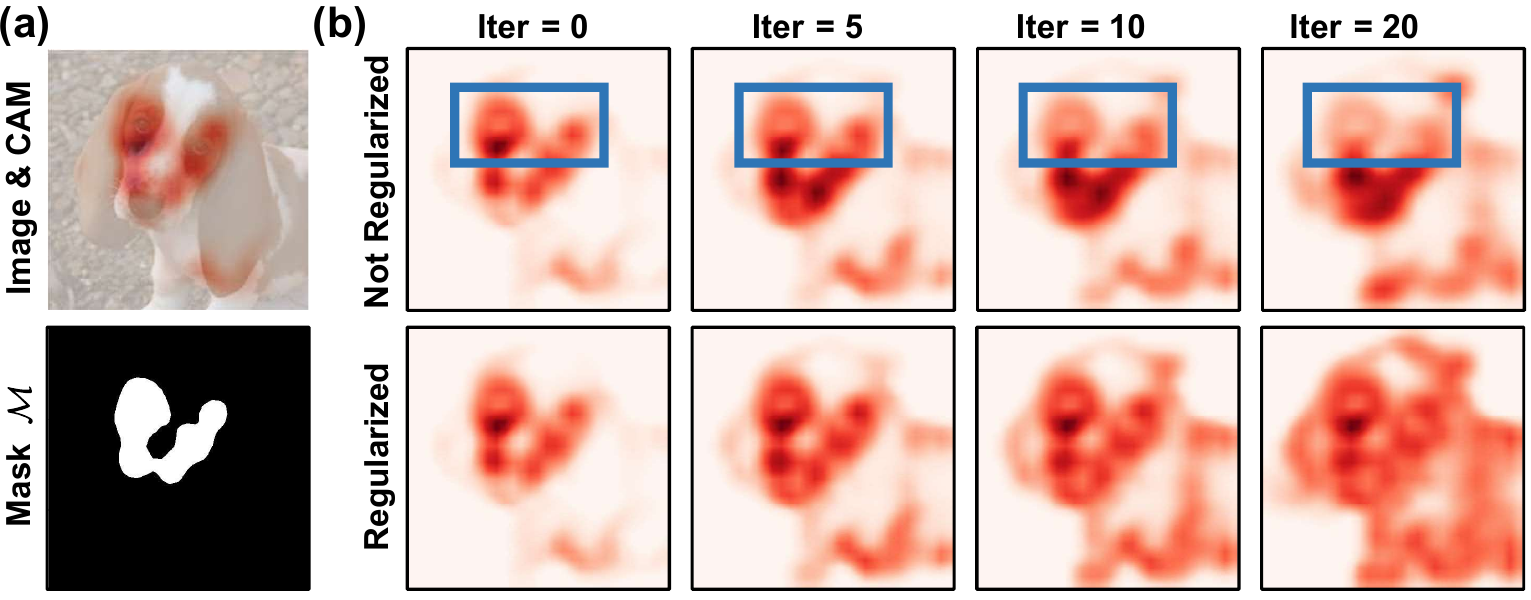}
\vspace{-1.6em}
\caption{\label{mask_ex} (a) An example image with its CAM and restricting mask $\mathcal{M}$. (b) The initial CAM, and CAMs after 5, 10 and 20 steps of adversarial climbing, with and without regularization.}
\vspace{-1.em}
\end{figure}

%% file: latex/tables/table_seed.tex
\begin{table}[tbp]
\renewcommand{\arraystretch}{0.95}
  \centering
  \caption{mIoU (\%) of the initial seed (Seed), the seed with CRF (+CRF), and the pseudo ground truth mask (Mask) on PASCAL VOC 2012 \textit{train} images.}
  \vspace{-0.7em}
    \begin{tabular}{l@{\hskip 0.3in}c@{\hskip 0.1in}c@{\hskip 0.08in}c}
     \Xhline{1pt}\\[-0.95em]
    Method  &  Seed  & + CRF & Mask \\
    \hline\hline \\[-0.9em]
    \multicolumn{4}{l}{Seed Refine with PSA~\cite{ahn2018learning}:} \\
    $\text{PSA}_{\text{~~CVPR '18}}$~\cite{ahn2018learning} & 48.0 & - & 61.0 \\
    $\text{Mixup-CAM}_{\text{~~BMVC '20}}$~\cite{chang2020mixup} & 50.1 & - & 61.9 \\
    $\text{Chang \textit{et al.}}_{\text{~~CVPR '20}}$~\cite{chang2020weakly} & 50.9 & 55.3 & 63.4 \\
    $\text{SEAM}_{\text{~~CVPR '20}}$~\cite{wang2020self}  & 55.4 & 56.8& 63.6 \\
    AdvCAM (Ours) & \textbf{55.6}  & \textbf{62.1} & \textbf{68.0}\\
    % $\text{CAM + BEM}_{\text{~~ECCV '20}}$~\cite{chenweakly}  & 50.4 & 67.2\\
    \hline\\[-0.9em]
    \multicolumn{4}{l}{Seed Refine with IRN~\cite{ahn2019weakly}:} \\
    $\text{IRN}_{\text{~~CVPR '19}}$~\cite{ahn2019weakly} & 48.8  & 54.3 & 66.3 \\
    $\text{MBMNet}_{\text{~~ACMMM '20}}$~\cite{liu2020weakly} & 50.2 & - & 66.8 \\
    $\text{CONTA}_{\text{~~NeurIPS '20}}$~\cite{zhang2020causal} & 48.8 & - & 67.9 \\

    AdvCAM (Ours) & \textbf{55.6}  & \textbf{62.1} & \textbf{69.9} \\
    \Xhline{1pt}
    \vspace{-2em}
    \end{tabular}%
  \label{table_seed}%
\end{table}%

%% file: latex/tables/table_weak_sem_pascal.tex
\begin{table}[t]
\renewcommand{\arraystretch}{0.95}
\centering
  \caption{Weakly supervised semantic segmentation performance on PASCAL VOC 2012 \textit{val} and \textit{test} images.}\label{table_semantic}
%   Level of supervision: $\mathcal{F}-$full, $\mathcal{I}-$image label, $\mathcal{B}-$box, $\mathcal{S}-$saliency.}  
\vspace{-0.7em}
\begin{threeparttable}
\begin{tabular}{l@{\hskip 0.3in}c@{\hskip 0.3in}cc}
    \Xhline{1pt}\\[-0.95em]
    Method  & Sup.& \textit{val} & \textit{test}\\
    \hline\hline 
    \\[-0.9em]
    
    \multicolumn{4}{l}{Supervision: Stronger than image labels} \\
    $\text{DeepLab}_{\text{~~TPAMI '17}}$~\cite{chen2017deeplab}  & $\mathcal{P}$  & 76.8  & 76.2 \\
    $\text{SDI}_{\text{~~CVPR '17}}$~\cite{khoreva2017simple}   & $\mathcal{B}$ & 69.4  & -  \\
    $\text{Song \textit{et al.}}_{\text{~~CVPR '19}}$~\cite{song2019box}   & $\mathcal{B}$ & 70.2  & - \\
    % $\text{Box2Seg}_{\text{~~ECCV '20}}$~\cite{kulharia12356box2seg}    & 76.4  & - \\
    \\[-0.9em]
\hline
    \\[-0.9em]
    \multicolumn{3}{l}{Supervision: Image-level tags}\\
%     MIL-FCN (ICLR '15) &   10K    & 25.7  & 24.9 \\
%     CCNN (ICCV '15) & 10K   & 35.3  & 35.6 \\
%     EM\_Adapt (ICCV '15) & 10K   & 38.2  & 39.6 \\
%     $\text{DCSM}_{\text{~~ECCV '16}}$~\cite{shimoda2016distinct} & 10K   & 44.1  & 45.1 \\
%     $\text{BFBP}_{\text{~~ECCV '16}}$~\cite{saleh2016built} & 10K   & 46.6  & 48.0 \\
%     $\text{SEC}_{\text{~~ECCV '16}}$~\cite{kolesnikov2016seed}    & 50.7  & 51.1 \\
% %     $\text{Saleh et al.}_{\text{~~TPAMI '17}}$~\cite{saleh2018incorporating} & 10K   & 50.9  & 52.6 \\
%     $\text{CBTS-cues}_{\text{~~CVPR '17}}$~\cite{roy2017combining}    & 52.8  & 53.7 \\
%     $\text{TPL}_{\text{~~ICCV '17}}$~\cite{kim2017two}   & 53.1  & 53.8 \\
%     $\text{AE\_PSL}_{\text{~~CVPR '17}}$~\cite{wei2017object}    & 55.0    & 55.7 \\
%     $\text{DCSP}_{\text{~~BMVC '17}}$~\cite{chaudhry2017discovering}  & 58.6 & 59.2\\
%     $\text{MEFF}_{\text{~~CVPR '18}}$~\cite{ge2018multi}  & - & 55.6\\
    % $\text{GAIN}_{\text{~~TPAMI '19}}$~\cite{li2019guided}    & 59.4  & 59.6 \\
    % $\text{MCOF}_{\text{~~CVPR '18}}$~\cite{wang2018weakly}    & 60.3  & 61.2 \\
    % $\text{DSRG}_{\text{~~CVPR '18}}$~\cite{huang2018weakly}   & 61.4   & 63.2 \\
    % $\text{AffinityNet}_{\text{~~CVPR '18}}$~\cite{ahn2018learning}    & 61.7  & 63.7 \\
    $\text{Li \textit{et al.}}_{\text{~~ICCV '19}}$~\cite{li2019attention}   &   $\mathcal{I}$, $\mathcal{S}$ & 62.1  & 63.0  \\
    % $\text{SeeNet}_{\text{~~NeurIPS '18}}$~\cite{hou2018self}  & $\mathcal{I}$, $\mathcal{S}$ & 63.1 & 62.8 \\
    $\text{FickleNet}_{\text{~~CVPR '19}}$~\cite{lee2019ficklenet}  & $\mathcal{I}$, $\mathcal{S}$ & 64.9 & 65.3\\
    $\text{Lee \textit{et al.}}_{\text{~~ICCV '19}}$~\cite{lee2019frame}   & $\mathcal{I}$, $\mathcal{S}$, $\mathcal{W}$  & 66.5  & 67.4  \\
    % $\text{OAA+}_{\text{~~ICCV '19}}$~\cite{JiangOAAICCV19}     & 65.2  & 66.4 \\
    $\text{CIAN}_{\text{~~AAAI '20}}$~\cite{fan2018cian}    & $\mathcal{I}$, $\mathcal{S}$ & 64.3  & 65.3 \\
    $\text{Zhang \textit{et al.}}_{\text{~~ECCV '20}}$~\cite{zhangsplitting}   & $\mathcal{I}$, $\mathcal{S}$  & 66.6  & 66.7  \\
    $\text{Sun \textit{et al.}}_{\text{~~ECCV '20}}$~\cite{sun2020mining}   & $\mathcal{I}$, $\mathcal{S}$  & 66.2  & 66.9  \\
    $\text{Fan \textit{et al.}}_{\text{~~ECCV '20}}$~\cite{fanemploying}   & $\mathcal{I}$, $\mathcal{S}$  & 67.2  & 66.7  \\
    $\text{Sun \textit{et al.}}_{\text{~~ECCV '20}}$~\cite{sun2020mining}   & $\mathcal{I}$, $\mathcal{S}$, $\mathcal{W}$  & 67.7  & 67.5  \\
    
    $\text{IRN}_{\text{~~CVPR '19}}$~\cite{ahn2019weakly}  &  $\mathcal{I}$ & 63.5 & 64.8 \\
    $\text{SSDD}_{\text{~~ICCV '19}}$~\cite{Shimoda_2019_ICCV}    & $\mathcal{I}$   & 64.9  & 65.5\\
    $\text{SEAM}_{\text{~~CVPR '20}}$~\cite{wang2020self}    & $\mathcal{I}$ & 64.5  & 65.7 \\
    % $\text{Mixup-CAM}_{\text{~~BMVC '20}}$~\cite{chang2020mixup}   & $\mathcal{I}$  & 65.6  & -\\
    $\text{Chen \textit{et al.}}_{\text{~~ECCV '20}}$~\cite{chenweakly}   &   $\mathcal{I}$ & 65.7  & 66.6  \\

    $\text{Chang \textit{et al.}}_{\text{~~CVPR '20}}$~\cite{chang2020weakly}   & $\mathcal{I}$  & 66.1  & 65.9\\
    % $\text{RRM}_{\text{~~AAAI '20}}$~\cite{zhang2019reliability}     & 66.3  & 66.5   \\
    
    $\text{CONTA}_{\text{~~NeurIPS '20}}$~\cite{zhang2020causal}   & $\mathcal{I}$  & 66.1  & 66.7  \\

    AdvCAM (Ours) & $\mathcal{I}$ & \textbf{68.1} & \textbf{68.0}  \\
    % \\[-0.9em]
    \Xhline{1pt}
    
    \end{tabular}%
    \begin{tablenotes}
  \footnotesize
\item $\mathcal{P}-$pixel-level mask, $\mathcal{I}-$image class, $\mathcal{B}-$box, $\mathcal{S}-$saliency, $\mathcal{W}-$web\\
\scriptsize
% $^*$ Anonymous result link can be found \href{http://host.robots.ox.ac.uk:8080/anonymous/TQSZTQ.html}{here}.
% $^*$\url{http://host.robots.ox.ac.uk:8080/anonymous/NE2KAJ.html}
        \end{tablenotes}
     \end{threeparttable}
    % \end{adjustbox}
    \vspace{-1em}

      \end{table}

%% file: latex/tables/table_semi.tex
\begin{table}[tbp]
\renewcommand{\arraystretch}{0.95}
  \centering  \caption{Comparison of semi-supervised semantic segmentation methods on the PASCAL VOC 2012 \textit{val} and \textit{test} images.}
%   \resizebox{0.48\textwidth}{!}{
\vspace{-0.7em}
\begin{threeparttable}
    \begin{tabular}{l@{\hskip 0.1in}c@{\hskip 0.15in}cc}
    \Xhline{1pt}\\[-0.95em]
    Method & Training set & \textit{val} & \textit{test}    \\
    \hline\hline\\[-0.95em]
    % DeepLab~\cite{chen2014semantic} & 1.4K strong & 62.5\\
    % \hline\\[-0.95em]
    WSSL$^{\dagger}$~\cite{papandreou2015weakly}  & 1.5K $\mathcal{P}$ + 9.1K $\mathcal{I}$ & 64.6 & 66.2 \\
    % GAIN~\cite{li2018tell}  & 1.5K $\mathcal{F}$ + 9.1K $\mathcal{I}$ & 60.5 & 62.1 \\
    MDC$^{\dagger}$~\cite{wei2018revisiting}  & 1.5K $\mathcal{P}$ + 9.1K $\mathcal{I}$ & 65.7 & 67.6\\
    
    % DSRG ~\cite{huang2018weakly} (baseline) & 1.4K strong + 9.1K weak& 64.3\\
    Souly \textit{et al.}$^{\dagger}$~\cite{souly2017semi} & 1.5K $\mathcal{P}$ + 9.1K $\mathcal{I}$ &   65.8  & - \\
    FickleNet$^{\dagger}$~\cite{lee2019ficklenet} & 1.5K $\mathcal{P}$ + 9.1K $\mathcal{I}$ &   65.8  & - \\
    Song \textit{et al.}~\cite{song2019box}& 1.5K $\mathcal{P}$ + 9.1K $\mathcal{B}$ &   71.6 & -  \\

    Luo \textit{et al.}~\cite{luosemi}& 1.5K $\mathcal{P}$ + 9.1K $\mathcal{I}$ &   76.6 & -  \\
    CCT~\cite{ouali2020semi} (baseline)& 1.5K $\mathcal{P}$ + 9.1K $\mathcal{I}$  &   73.2 & -  \\
    AdvCAM (Ours) & 1.5K $\mathcal{P}$ + 9.1K $\mathcal{I}$ &   \textbf{77.8}  &  \textbf{76.9}\\
    % advCAM (Ours) & 1.5K $\mathcal{F}$ + 9.1K $\mathcal{I}$ &   \textbf{76.9}  &  \\

    % \hline \\[-0.95em]
    % DeepLab~\cite{chen2014semantic} & 10.6K strong & 67.6  \\
    \Xhline{1pt}
    \end{tabular}%
    \begin{tablenotes}
  \footnotesize
\item $\mathcal{P}-$pixel-level mask, $\mathcal{I}-$image class label, $\mathcal{B}-$box, $^{\dagger}-$ VGG backbone \\
% $^*$ Anonymous result link can be found \href{http://host.robots.ox.ac.uk:8080/anonymous/TQSZTQ.html}{here}.
\scriptsize
% $^*$\url{http://host.robots.ox.ac.uk:8080/anonymous/TQSZTQ.html}
        \end{tablenotes}
     \end{threeparttable}
  \label{tabsemi}
%   }%
    % \vspace*{-0.55\baselineskip}
    \vspace{-1.3em}

\end{table}%

%% file: latex/Figures/figure_seg_samples.tex
\begin{figure*}[t]
\centering
\includegraphics[width=0.96\linewidth]{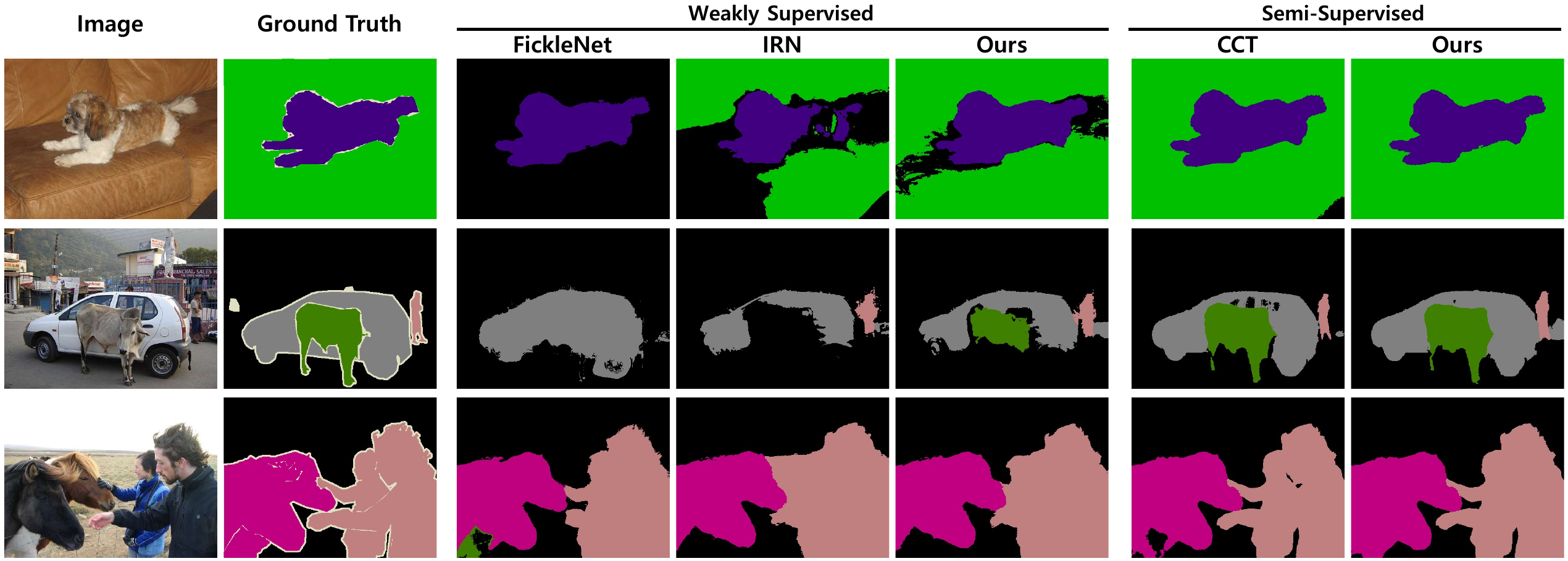}
\vspace{-.7em}
\caption{\label{segsample} Examples of predicted semantic masks for PASCAL VOC \textit{val} images in weakly and semi-supervised manner.}
% Effectiveness of regularization technique. (\textit{Left}) Seed quality on mIoU (\%). (\textit{Right}) 1-precision (\%)}
\vspace{-1em}
\end{figure*}

%% file: latex/Figures/figure_each_iter.tex
\begin{figure*}[t]
\centering
\includegraphics[width=0.96\linewidth]{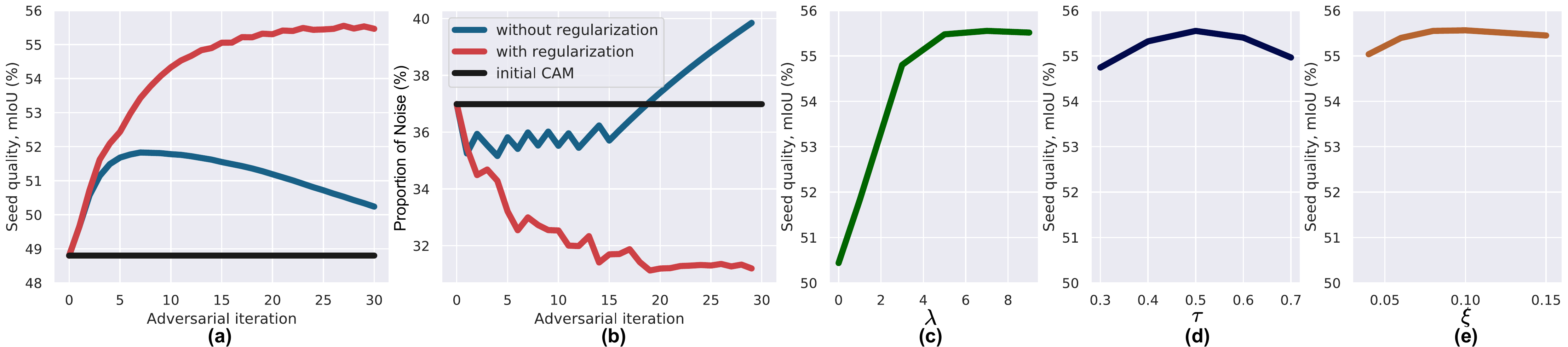}
\vspace{-.7em}
\caption{\label{eachiter} 
Effect of adversarial climbing and regularization on (a) the seed quality and (b) the proportion of noise. (c) Effect of the regularization coefficient $\lambda$. (d) Effect of the masking threshold $\tau$. (d) Effect of the step size $\xi$.}
\vspace{-1em}
\end{figure*}

%% file: latex/Figures/figure_each_iter_examples.tex
\begin{figure*}[t]
\centering
\vspace{-0.2em}
\includegraphics[width=\linewidth]{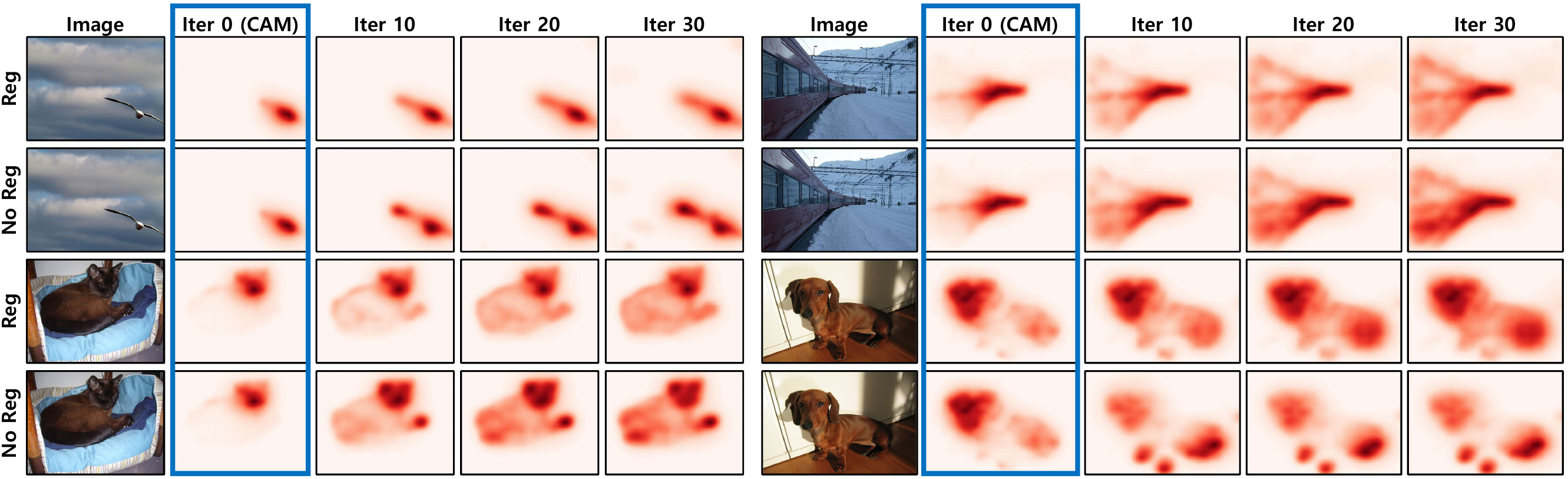}
\vspace{-2em}
\caption{\label{eachiter_ex} Examples of initial CAMs (the blue boxes) and successive localization maps obtained from images manipulated by iterative adversarial climbing, with the regularization procedure (\textit{top}) and without (\textit{bottom}).}
\vspace{-1.em}
\end{figure*}

%% file: latex/tables/table_pseudo_gt.tex
\begin{table}[tbp]
\renewcommand{\arraystretch}{0.92}
  \centering
  \caption{Effects of AdvCAM on different methods of generating the initial seed: mIoU of the initial seed (Seed) and of the pseudo ground truth mask (Mask), for the PASCAL VOC 2012 training images.}
  \vspace{-0.7em}
    \begin{tabular}{l@{\hskip 0.25in}l@{\hskip 0.25in}l}
     \Xhline{1pt}\\[-0.95em]
    Method  & Seed  & Mask \\
    \hline\hline \\[-0.9em]
    $\text{Chang \textit{et al.}}$~\cite{chang2020weakly} & 50.9 & 63.4 \\
    + AdvCAM  &53.7$_{~\textcolor{red}{\scaleto{+2.8}{6pt}}}$ & 67.5$_{~\textcolor{red}{\scaleto{+4.1}{6pt}}}$  \\
    \hline \\[-0.9em]
    $\text{SEAM}$~\cite{wang2020self} & 55.4  & 63.6  \\
    + AdvCAM  & 58.6$_{~\textcolor{red}{\scaleto{+3.2}{6pt}}}$ & 67.2$_{~\textcolor{red}{\scaleto{+3.6}{6pt}}}$   \\
    \hline \\[-0.9em]
    $\text{IRN}$~\cite{ahn2018learning} & 48.8  & 66.3  \\
    + AdvCAM  & 55.6$_{~\textcolor{red}{\scaleto{+6.8}{6pt}}}$  & 69.9$_{~\textcolor{red}{\scaleto{+3.6}{6pt}}}$   \\
    % \hline\\[-0.95em]
    % advCAM (Ours)  & \xmark  & 36.3 & 13.1 \\
    % advCAM (Ours) & \textcolor{red}{\cmark} & - & - \\
    \Xhline{1pt}
    \vspace{-1.5em}
    \end{tabular}%
  \label{tab:baselines}%
\end{table}%

% \begin{table}[htbp]
% % \renewcommand{\arraystretch}{0.97}
%   \centering
%   \caption{Different baselines + advCAM on PASCAL VOC 2012 train images in mIoU(\%)}
%   \vspace{-0.7em}
%     \begin{tabular}{cccc}
%      \Xhline{1pt}\\[-0.95em]
%     advCAM  & Chang \textit{et al.}~\cite{chang2020weakly}  & SEAM~\cite{wang2020self} & IRN~\cite{ahn2019weakly}\\
%     \hline\hline \\[-0.9em]
%      -  & 50.9 & 63.4 \\
%     \checkmark  & 52.9$_{~\textcolor{red}{\scaleto{+2.0}{6pt}}}$ & 66.3$_{~\textcolor{red}{\scaleto{+2.9}{6pt}}}$  \\
%     \hline \\[-0.9em]
%     - & 55.4  & 63.6  \\
%     \checkmark  & 57.5$_{~\textcolor{red}{\scaleto{+2.1}{6pt}}}$ & 64.4$_{~\textcolor{red}{\scaleto{+0.8}{6pt}}}$   \\
%     \hline \\[-0.9em]
%     - & 48.8  & 66.3  \\
%     \checkmark  & 55.6$_{~\textcolor{red}{\scaleto{+6.8}{6pt}}}$  & 69.4$_{~\textcolor{red}{\scaleto{+3.1}{6pt}}}$   \\
%     % \hline\\[-0.95em]
%     % advCAM (Ours)  & \xmark  & 36.3 & 13.1 \\
%     % advCAM (Ours) & \textcolor{red}{\cmark} & - & - \\
%     \Xhline{1pt}
%     \end{tabular}%
%   \label{tab:baselines}%
% \end{table}%

%% file: latex/Figures/figure_tsne.tex
\begin{figure}[t]
\centering
\includegraphics[width=\linewidth]{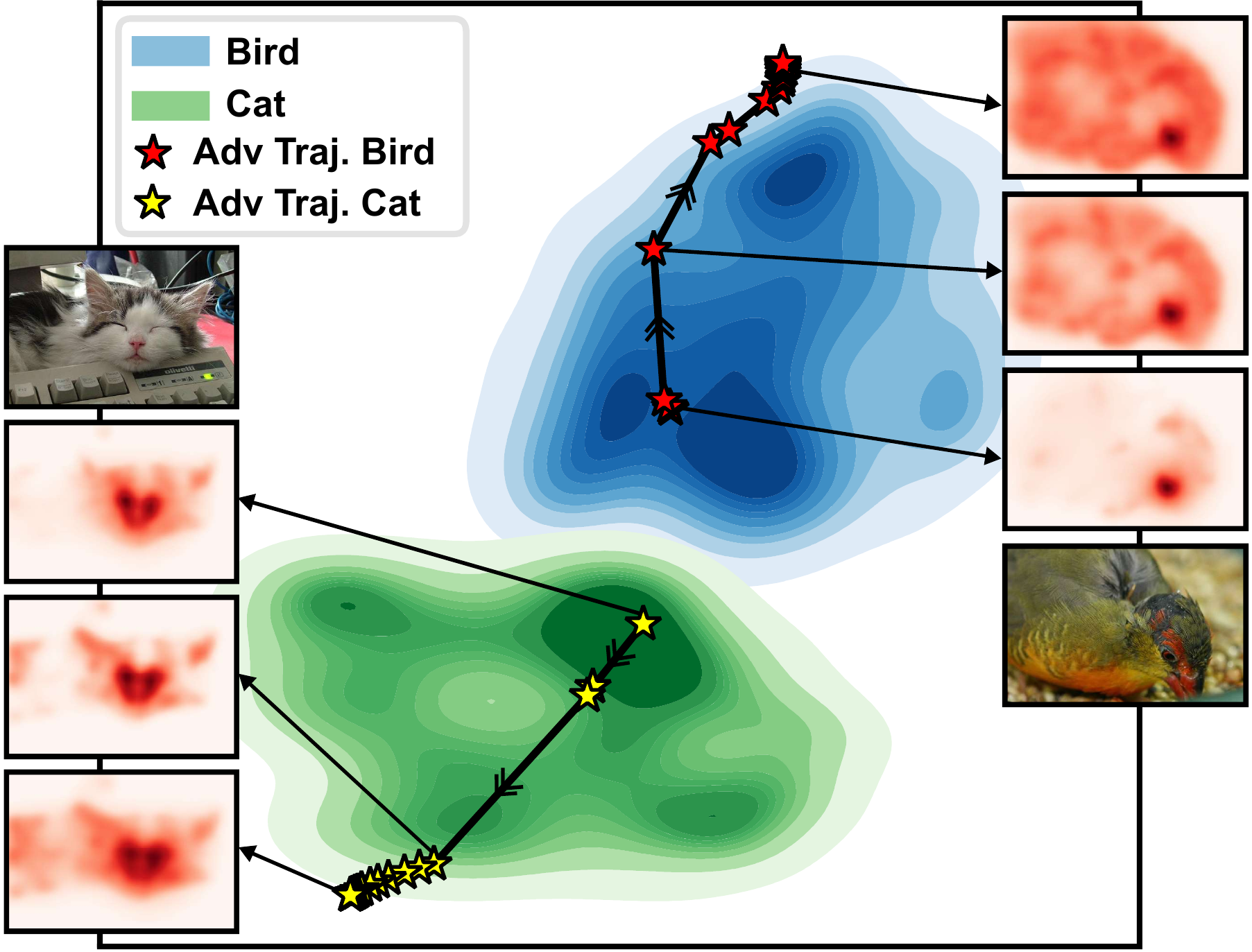}
% \vspace{-1.5em}
\caption{\label{fig_tsne} Feature manifold of images with ``bird" (blue) and ``cat" (green), and a trajectory of adversarial climbing for an image of each class. The dimensionality of the feature was reduced by t-SNE~\cite{maaten2008visualizing}.
}
\vspace{-1em}
\end{figure}

%% file: latex/Figures/fig_threshold.tex
\begin{figure*}[t]
\centering
\includegraphics[width=0.8\linewidth]{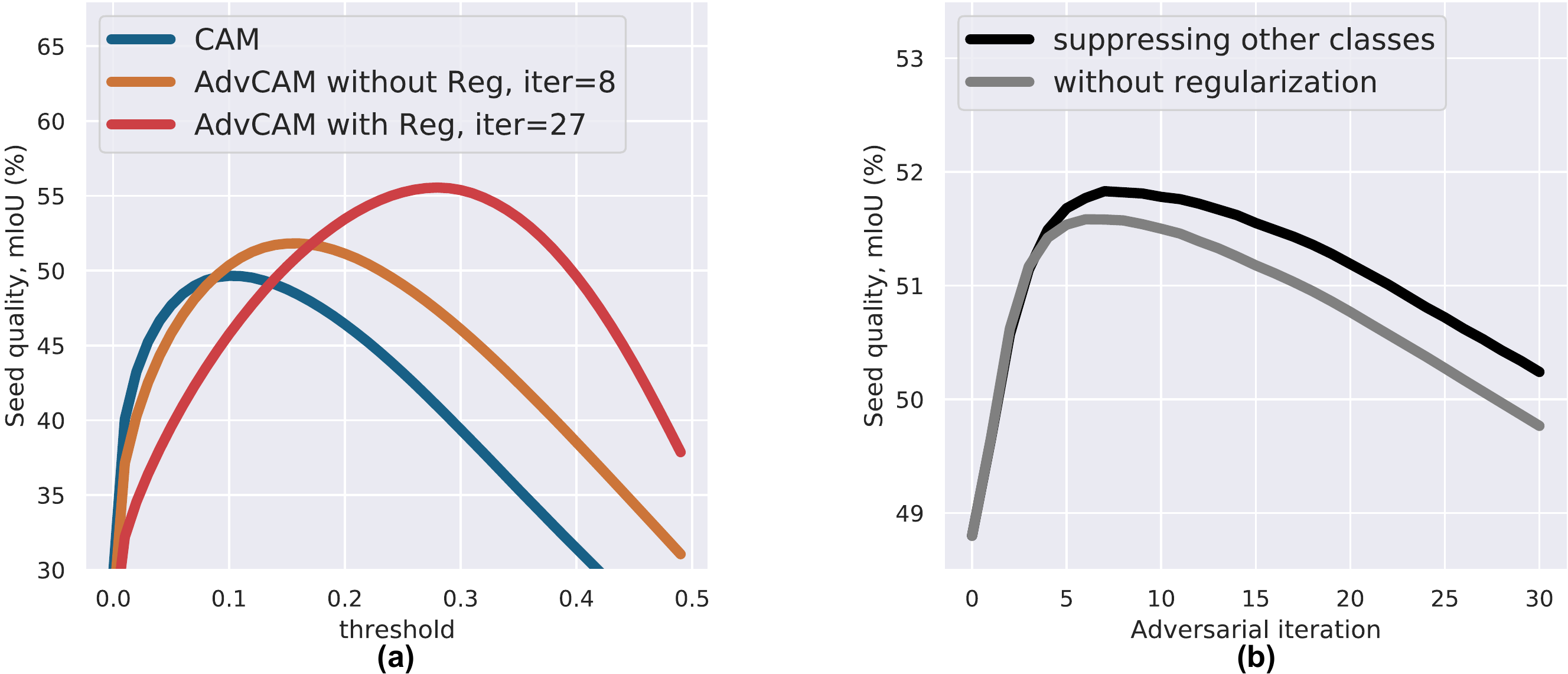}
\vspace{-1em}
\caption{\label{appendix_eachiter_ex} (a) Threshold analysis. (b) Effect of suppressing other classes.}
\vspace{-1em}
\end{figure*}

%% file: latex/tables/table_per_class.tex
\renewcommand{\tabcolsep}{2pt}

\begin{table*}[t]
  \caption{Comparison of per-class mIoU scores.}
  \vspace{-1em}
% \resizebox{\textwidth}{!}{%
  \centering
  \begin{adjustbox}{max width=\textwidth}
    \begin{tabular}{lccccccccccccccccccccc|c}
    
        \Xhline{1pt}

        \\[-0.95em]
     & bkg& aero  & bike  & bird  & boat  & bottle & bus   & car   & cat   & chair & cow   & table & dog   & horse & motor & person & plant & sheep & sofa  & train & tv  ~ & mIOU \\
   
    \hline
    \hline
    \\[-0.9em]
    % \multicolumn{22}{l}{Results on validation images (without CRF):}\\
    % MDC &   88.5    &   77.9    &   32.5    &   68.3    &   56.7    &    59.9   &    64.2   &   70.6    &   73.2    &    17.0   &   63.7    &   12.2    &    69.8   &    62.7   &   67.5    &    68.5   &   32.9    &   68.1    &   24.8    &    70.3   &    49.5  ~ &  57.1\\
    % FickleNet &   87.6    &   71.6    &   31.4    &   73.3    &   46.8    &    58.5   &    78.7   &   71.4    &   77.8    &    24.4   &   64.2    &   41.2    &    75.1   &    65.4   &   65.3    &    68.9   &   42.5    &   69.8    &   33.7    &    53.3   &    53.2  ~ &  59.7\\
   \multicolumn{22}{l}{Results on PASCAL VOC 2012 validation images:}\\

    AdvCAM (Ours, weak) &   90.0    &   79.8    &   34.1    &   82.6    &   63.3   &   70.5    &    89.4  &   76.0   &    87.3   &  31.4     &   81.3    &   33.1    &    82.5 &    80.8   &   74.0    &  72.9 &  50.3   &    82.3  & 42.2  &  74.1 &   52.9  & 68.1\\
    AdvCAM (Ours, semi) &   94.4    &   91.7    &   65.6    &   89.1    &   72.4   &   72.8    &    93.4  &   86.0   &    90.4   &  37.5     &   90.6    &   58.6    &    84.5 &    88.9   &   83.3    &  84.9  &  62.0   &    81.6  & 49.5  &  85.9 &   71.8  & 77.8\\
    \hline\\[-0.9em]
    \multicolumn{22}{l}{Results on PASCAL VOC 2012 test images:}\\
    % PSA~\cite{ahn2018learning}&   89.1    &   70.6    &   31.6    &   77.2    &   42.2   &   68.9    &    79.1  &    66.5   &    74.9   &  29.6     &   68.7    &   56.1    &    82.1 &    64.8   &   78.6    &  73.5 &  50.8   & 70.7  & 47.7  &  63.9 &   51.1~ &  63.7\\
    % FickleNet~\cite{lee2019ficklenet} &   90.3    &   77.0    &   35.2    &   76.0    &   54.2   &   64.3    &    76.6  &    76.1   &    80.2   &  25.7     &   68.6    &   50.2    &    74.6&    71.8   &   78.3    &  69.5 &  53.8   &    76.5  & 41.8  &  70.0 &   54.2~ &  65.0\\
    % SSDD~\cite{Shimoda_2019_ICCV} &   89.0    &   62.5    &   28.9    &   83.7    &   52.9   &   59.5    &    77.6  &    73.7   &    87.0   &  34.0     &   83.7    &   47.6    &    84.1 &    77.0   &   73.9    &  69.6 &  29.8   &    84.0  & 43.2  &  68.0 &   53.4~ &  64.9\\
    % Lee \textit{et al.}~\cite{lee2019frame} &   91.2    &   84.2    &   37.9    &   81.6    &   53.8   &   70.6    &    79.2  &    75.6   &    82.3   &  29.3     &   76.2    &   35.6    &    81.4&    80.5   &   79.9    &  76.8 &  44.7   &    83.0  & 36.1  &  74.1 &   60.3  &  67.4\\
    AdvCAM (Ours, weak) &   90.1  &   81.2    &   33.6    &   80.4    &   52.4   &   66.6    &    87.1  &   80.5   &    87.2   &  28.9     &   80.1    &   38.5    &    84.0 &  83.0 &  79.5   &   71.9    &  47.5 &  80.8   &    59.1  & 65.4  &  49.7  & 68.0\\
    AdvCAM (Ours, semi) &   94.3  &   93.6    &   65.7    &   90.3    &   54.2   &   74.4    &    91.7  &   85.6   &    91.7  &  28.2     &   88.1    &   67.4    &    86.2 &    88.5   &   89.4    &  82.6 &  62.2   &    87.2  & 47.6  &  80.5 &   65.3  & 76.9\\
        \Xhline{1pt}
    \vspace{-2em}
    \end{tabular}%
  \end{adjustbox}%
  \label{class-specific-results}%
\end{table*}%

%% file: latex/Figures/seg_ex.tex
\begin{figure*}[t]
\centering
\includegraphics[width=\linewidth]{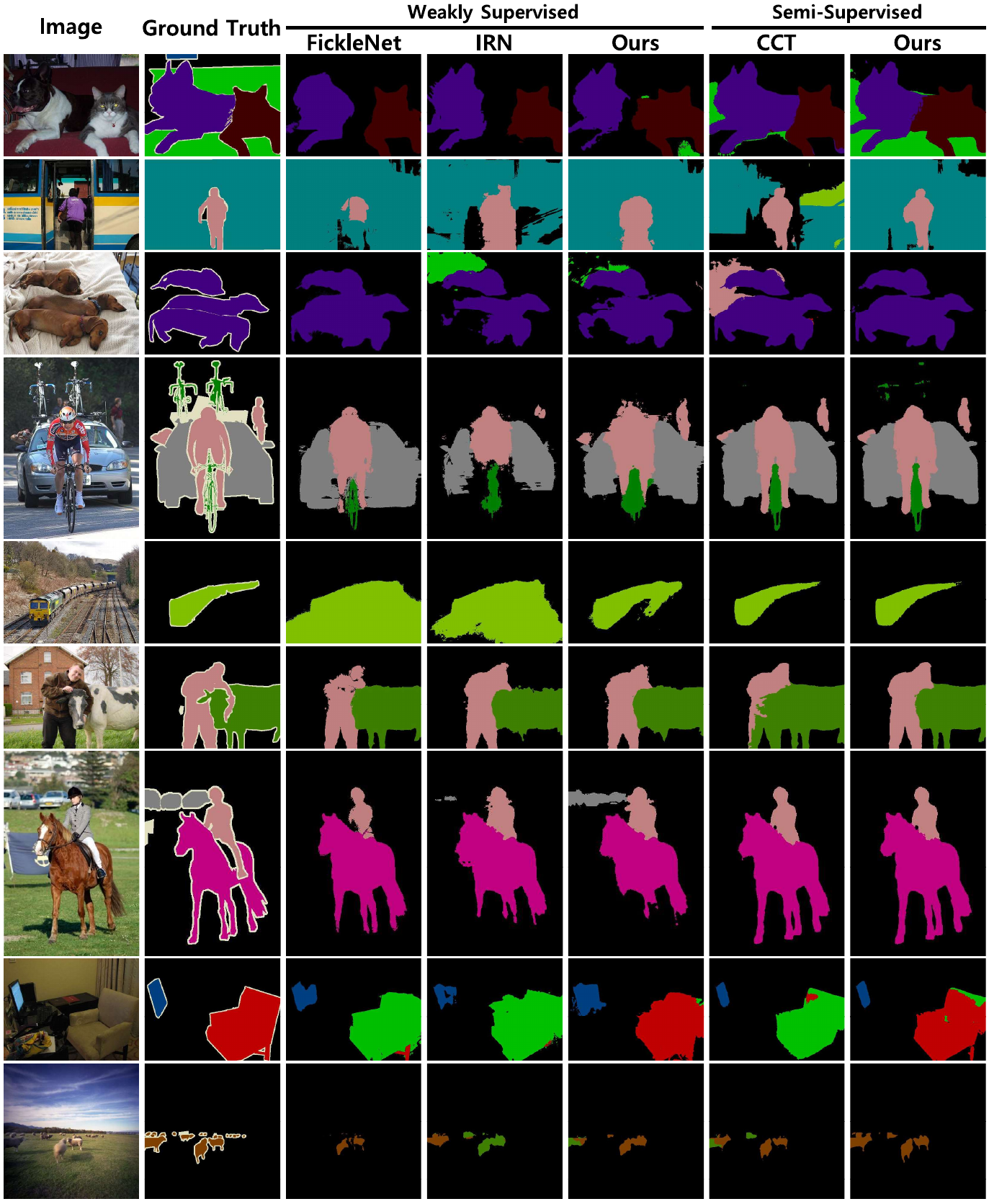}
\caption{\label{appendix_segsample} Examples of predicted semantic masks for PASCAL VOC \textit{val} images in weakly and semi-superivsed manner.}
% Effectiveness of regularization technique. (\textit{Left}) Seed quality on mIoU (\%). (\textit{Right}) 1-precision (\%)}
\end{figure*}

%% file: latex/Figures/seed_ex.tex
\begin{figure*}[t]
\centering
\includegraphics[width=\linewidth]{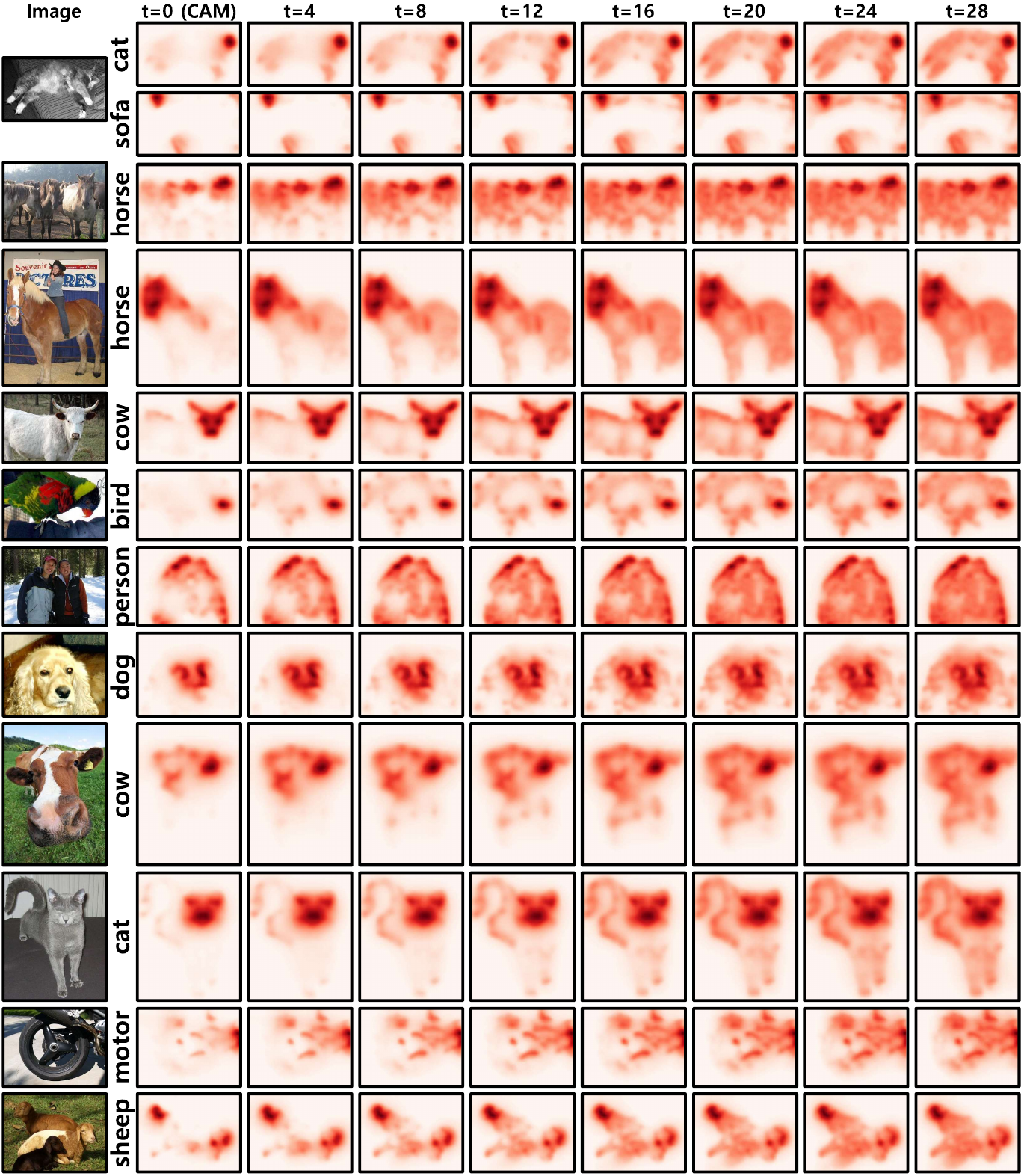}
\caption{\label{appendix_eachiter_ex_successive} Examples of initial CAMs and successive localization maps obtained from images manipulated by iterative adversarial climbing.}
% Effectiveness of regularization technique. (\textit{Left}) Seed quality on mIoU (\%). (\textit{Right}) 1-precision (\%)}
\end{figure*}

%% file: 00main.bbl
\begin{thebibliography}{10}\itemsep=-1pt

\bibitem{ahn2019weakly}
Jiwoon Ahn, Sunghyun Cho, and Suha Kwak.
\newblock Weakly supervised learning of instance segmentation with inter-pixel
  relations.
\newblock In {\em CVPR}, 2019.

\bibitem{ahn2018learning}
Jiwoon Ahn and Suha Kwak.
\newblock Learning pixel-level semantic affinity with image-level supervision
  for weakly supervised semantic segmentation.
\newblock In {\em CVPR}, 2018.

\bibitem{arnab2018robustness}
Anurag Arnab, Ondrej Miksik, and Philip~HS Torr.
\newblock On the robustness of semantic segmentation models to adversarial
  attacks.
\newblock In {\em CVPR}, 2018.

\bibitem{bearman2016s}
Amy Bearman, Olga Russakovsky, Vittorio Ferrari, and Li Fei-Fei.
\newblock What’s the point: Semantic segmentation with point supervision.
\newblock In {\em ECCV}, 2016.

\bibitem{chang2020mixup}
Yu-Ting Chang, Qiaosong Wang, Wei-Chih Hung, Robinson Piramuthu, Yi-Hsuan Tsai,
  and Ming-Hsuan Yang.
\newblock Mixup-cam: Weakly-supervised semantic segmentation via uncertainty
  regularization.
\newblock In {\em BMVC}, 2020.

\bibitem{chang2020weakly}
Yu-Ting Chang, Qiaosong Wang, Wei-Chih Hung, Robinson Piramuthu, Yi-Hsuan Tsai,
  and Ming-Hsuan Yang.
\newblock Weakly-supervised semantic segmentation via sub-category exploration.
\newblock In {\em CVPR}, 2020.

\bibitem{chenweakly}
Liyi Chen, Weiwei Wu, Chenchen Fu, Xiao Han, and Yuntao Zhang.
\newblock Weakly supervised semantic segmentation with boundary exploration.
\newblock In {\em ECCV}, 2020.

\bibitem{chen2017deeplab}
Liang-Chieh Chen, George Papandreou, Iasonas Kokkinos, Kevin Murphy, and Alan~L
  Yuille.
\newblock Deeplab: Semantic image segmentation with deep convolutional nets,
  atrous convolution, and fully connected crfs.
\newblock {\em IEEE TPAMI}, 2017.

\bibitem{cordts2016cityscapes}
Marius Cordts, Mohamed Omran, Sebastian Ramos, Timo Rehfeld, Markus Enzweiler,
  Rodrigo Benenson, Uwe Franke, Stefan Roth, and Bernt Schiele.
\newblock The cityscapes dataset for semantic urban scene understanding.
\newblock In {\em CVPR}, 2016.

\bibitem{deng2009imagenet}
Jia Deng, Wei Dong, Richard Socher, Li-Jia Li, Kai Li, and Li Fei-Fei.
\newblock Imagenet: A large-scale hierarchical image database.
\newblock In {\em CVPR}, 2009.

\bibitem{dombrowski2019explanations}
Ann-Kathrin Dombrowski, Maximillian Alber, Christopher Anders, Marcel
  Ackermann, Klaus-Robert M{\"u}ller, and Pan Kessel.
\newblock Explanations can be manipulated and geometry is to blame.
\newblock In {\em NeurIPS}, 2019.

\bibitem{everingham2010pascal}
Mark Everingham, Luc Van~Gool, Christopher~KI Williams, John Winn, and Andrew
  Zisserman.
\newblock The pascal visual object classes (voc) challenge.
\newblock {\em IJCV}, 2010.

\bibitem{fan2018cian}
Junsong Fan, Zhaoxiang Zhang, and Tieniu Tan.
\newblock Cian: Cross-image affinity net for weakly supervised semantic
  segmentation.
\newblock {\em AAAI}, 2020.

\bibitem{fanemploying}
Junsong Fan, Zhaoxiang Zhang, and Tieniu Tan.
\newblock Employing multi-estimations for weakly-supervised semantic
  segmentation.
\newblock In {\em ECCV}, 2020.

\bibitem{goodfellow2014generative}
Ian Goodfellow, Jean Pouget-Abadie, Mehdi Mirza, Bing Xu, David Warde-Farley,
  Sherjil Ozair, Aaron Courville, and Yoshua Bengio.
\newblock Generative adversarial nets.
\newblock In {\em NeurIPS}, 2014.

\bibitem{goodfellow2014explaining}
Ian~J Goodfellow, Jonathon Shlens, and Christian Szegedy.
\newblock Explaining and harnessing adversarial examples.
\newblock In {\em ICLR}, 2015.

\bibitem{hariharan2011semantic}
Bharath Hariharan, Pablo Arbel{\'a}ez, Lubomir Bourdev, Subhransu Maji, and
  Jitendra Malik.
\newblock Semantic contours from inverse detectors.
\newblock In {\em ICCV}, 2011.

\bibitem{he2016deep}
Kaiming He, Xiangyu Zhang, Shaoqing Ren, and Jian Sun.
\newblock Deep residual learning for image recognition.
\newblock In {\em CVPR}, 2016.

\bibitem{heo2019fooling}
Juyeon Heo, Sunghwan Joo, and Taesup Moon.
\newblock Fooling neural network interpretations via adversarial model
  manipulation.
\newblock In {\em NeurIPS}, 2019.

\bibitem{hong2017weakly}
Seunghoon Hong, Donghun Yeo, Suha Kwak, Honglak Lee, and Bohyung Han.
\newblock Weakly supervised semantic segmentation using web-crawled videos.
\newblock In {\em CVPR}, 2017.

\bibitem{hou2018self}
Qibin Hou, PengTao Jiang, Yunchao Wei, and Ming-Ming Cheng.
\newblock Self-erasing network for integral object attention.
\newblock In {\em NeurIPS}, 2018.

\bibitem{huang2019ccnet}
Zilong Huang, Xinggang Wang, Lichao Huang, Chang Huang, Yunchao Wei, and Wenyu
  Liu.
\newblock Ccnet: Criss-cross attention for semantic segmentation.
\newblock In {\em ICCV}, 2019.

\bibitem{huang2018weakly}
Zilong Huang, Xinggang Wang, Jiasi Wang, Wenyu Liu, and Jingdong Wang.
\newblock Weakly-supervised semantic segmentation network with deep seeded
  region growing.
\newblock In {\em CVPR}, 2018.

\bibitem{hung2019adversarial}
Wei~Chih Hung, Yi~Hsuan Tsai, Yan~Ting Liou, Yen~Yu Lin, and Ming~Hsuan Yang.
\newblock Adversarial learning for semi-supervised semantic segmentation.
\newblock In {\em BMVC}, 2018.

\bibitem{ilyas2019adversarial}
Andrew Ilyas, Shibani Santurkar, Dimitris Tsipras, Logan Engstrom, Brandon
  Tran, and Aleksander Madry.
\newblock Adversarial examples are not bugs, they are features.
\newblock In {\em NeurIPS}, 2019.

\bibitem{khoreva2017simple}
Anna Khoreva, Rodrigo Benenson, Jan Hosang, Matthias Hein, and Bernt Schiele.
\newblock Simple does it: Weakly supervised instance and semantic segmentation.
\newblock In {\em CVPR}, 2017.

\bibitem{kolesnikov2016seed}
Alexander Kolesnikov and Christoph~H Lampert.
\newblock Seed, expand and constrain: Three principles for weakly-supervised
  image segmentation.
\newblock In {\em ECCV}, 2016.

\bibitem{kurakin2016adversarial}
Alexey Kurakin, Ian Goodfellow, and Samy Bengio.
\newblock Adversarial machine learning at scale.
\newblock In {\em ICLR}, 2017.

\bibitem{lee2019ficklenet}
Jungbeom Lee, Eunji Kim, Sungmin Lee, Jangho Lee, and Sungroh Yoon.
\newblock Ficklenet: Weakly and semi-supervised semantic image segmentation
  using stochastic inference.
\newblock In {\em CVPR}, 2019.

\bibitem{lee2019frame}
Jungbeom Lee, Eunji Kim, Sungmin Lee, Jangho Lee, and Sungroh Yoon.
\newblock Frame-to-frame aggregation of active regions in web videos for weakly
  supervised semantic segmentation.
\newblock In {\em ICCV}, 2019.

\bibitem{lee2018robust}
Sungmin Lee, Jangho Lee, Jungbeom Lee, Chul-Kee Park, and Sungroh Yoon.
\newblock Robust tumor localization with pyramid grad-cam.
\newblock {\em arXiv preprint arXiv:1805.11393}, 2018.

\bibitem{li2018tell}
Kunpeng Li, Ziyan Wu, Kuan-Chuan Peng, Jan Ernst, and Yun Fu.
\newblock Tell me where to look: Guided attention inference network.
\newblock In {\em CVPR}, 2018.

\bibitem{li2019attention}
Kunpeng Li, Yulun Zhang, Kai Li, Yuanyuan Li, and Yun Fu.
\newblock Attention bridging network for knowledge transfer.
\newblock In {\em ICCV}, 2019.

\bibitem{li2014secrets}
Yin Li, Xiaodi Hou, Christof Koch, James~M Rehg, and Alan~L Yuille.
\newblock The secrets of salient object segmentation.
\newblock In {\em CVPR}, 2014.

\bibitem{liu2010learning}
Tie Liu, Zejian Yuan, Jian Sun, Jingdong Wang, Nanning Zheng, Xiaoou Tang, and
  Heung-Yeung Shum.
\newblock Learning to detect a salient object.
\newblock {\em TPAMI}, 2010.

\bibitem{liu2020weakly}
Weide Liu, Chi Zhang, Guosheng Lin, Tzu-Yi HUNG, and Chunyan Miao.
\newblock Weakly supervised segmentation with maximum bipartite graph matching.
\newblock In {\em ACMMM}, 2020.

\bibitem{luosemi}
Wenfeng Luo and Meng Yang.
\newblock Semi-supervised semantic segmentation via strong-weak dual-branch
  network.
\newblock In {\em ECCV}, 2020.

\bibitem{maaten2008visualizing}
Laurens van~der Maaten and Geoffrey Hinton.
\newblock Visualizing data using t-sne.
\newblock {\em Journal of machine learning research}, 2008.

\bibitem{moosavi2016deepfool}
Seyed-Mohsen Moosavi-Dezfooli, Alhussein Fawzi, and Pascal Frossard.
\newblock Deepfool: a simple and accurate method to fool deep neural networks.
\newblock In {\em CVPR}, 2016.

\bibitem{moosavi2019robustness}
Seyed-Mohsen Moosavi-Dezfooli, Alhussein Fawzi, Jonathan Uesato, and Pascal
  Frossard.
\newblock Robustness via curvature regularization, and vice versa.
\newblock In {\em CVPR}, 2019.

\bibitem{ouali2020semi}
Yassine Ouali, C{\'e}line Hudelot, and Myriam Tami.
\newblock Semi-supervised semantic segmentation with cross-consistency
  training.
\newblock In {\em CVPR}, 2020.

\bibitem{papandreou2015weakly}
George Papandreou, Liang-Chieh Chen, Kevin Murphy, and Alan~L Yuille.
\newblock Weakly-and semi-supervised learning of a dcnn for semantic image
  segmentation.
\newblock In {\em ICCV}, 2015.

\bibitem{qin2019adversarial}
Chongli Qin, James Martens, Sven Gowal, Dilip Krishnan, Krishnamurthy
  Dvijotham, Alhussein Fawzi, Soham De, Robert Stanforth, and Pushmeet Kohli.
\newblock Adversarial robustness through local linearization.
\newblock In {\em NeurIPS}, 2019.

\bibitem{santurkar2019image}
Shibani Santurkar, Andrew Ilyas, Dimitris Tsipras, Logan Engstrom, Brandon
  Tran, and Aleksander Madry.
\newblock Image synthesis with a single (robust) classifier.
\newblock In {\em NeurIPS}, 2019.

\bibitem{selvaraju2017grad}
Ramprasaath~R Selvaraju, Michael Cogswell, Abhishek Das, Ramakrishna Vedantam,
  Devi Parikh, and Dhruv Batra.
\newblock Grad-cam: Visual explanations from deep networks via gradient-based
  localization.
\newblock In {\em ICCV}, 2017.

\bibitem{shen2018bootstrapping}
Tong Shen, Guosheng Lin, Chunhua Shen, and Ian Reid.
\newblock Bootstrapping the performance of webly supervised semantic
  segmentation.
\newblock In {\em CVPR}, 2018.

\bibitem{Shimoda_2019_ICCV}
Wataru Shimoda and Keiji Yanai.
\newblock Self-supervised difference detection for weakly-supervised semantic
  segmentation.
\newblock In {\em ICCV}, 2019.

\bibitem{simonyan2014very}
Karen Simonyan and Andrew Zisserman.
\newblock Very deep convolutional networks for large-scale image recognition.
\newblock {\em arXiv preprint arXiv:1409.1556}, 2014.

\bibitem{singh2017hide}
Krishna~Kumar Singh and Yong~Jae Lee.
\newblock Hide-and-seek: Forcing a network to be meticulous for
  weakly-supervised object and action localization.
\newblock In {\em ICCV}, 2017.

\bibitem{song2019box}
Chunfeng Song, Yan Huang, Wanli Ouyang, and Liang Wang.
\newblock Box-driven class-wise region masking and filling rate guided loss for
  weakly supervised semantic segmentation.
\newblock In {\em CVPR}, 2019.

\bibitem{souly2017semi}
Nasim Souly, Concetto Spampinato, and Mubarak Shah.
\newblock Semi supervised semantic segmentation using generative adversarial
  network.
\newblock In {\em ICCV}, 2017.

\bibitem{sun2020mining}
Guolei Sun, Wenguan Wang, Jifeng Dai, and Luc Van~Gool.
\newblock Mining cross-image semantics for weakly supervised semantic
  segmentation.
\newblock In {\em ECCV}, 2020.

\bibitem{tang2018normalized}
Meng Tang, Abdelaziz Djelouah, Federico Perazzi, Yuri Boykov, and Christopher
  Schroers.
\newblock Normalized cut loss for weakly-supervised cnn segmentation.
\newblock In {\em CVPR}, 2018.

\bibitem{tsipras2018robustness}
Dimitris Tsipras, Shibani Santurkar, Logan Engstrom, Alexander Turner, and
  Aleksander Madry.
\newblock Robustness may be at odds with accuracy.
\newblock {\em ICLR}, 2019.

\bibitem{wang2020self}
Yude Wang, Jie Zhang, Meina Kan, Shiguang Shan, and Xilin Chen.
\newblock Self-supervised equivariant attention mechanism for weakly supervised
  semantic segmentation.
\newblock In {\em CVPR}, 2020.

\bibitem{wei2017object}
Yunchao Wei, Jiashi Feng, Xiaodan Liang, Ming-Ming Cheng, Yao Zhao, and
  Shuicheng Yan.
\newblock Object region mining with adversarial erasing: A simple
  classification to semantic segmentation approach.
\newblock In {\em CVPR}, 2017.

\bibitem{wei2018revisiting}
Yunchao Wei, Huaxin Xiao, Honghui Shi, Zequn Jie, Jiashi Feng, and Thomas~S
  Huang.
\newblock Revisiting dilated convolution: A simple approach for weakly-and
  semi-supervised semantic segmentation.
\newblock In {\em CVPR}, 2018.

\bibitem{wu2019wider}
Zifeng Wu, Chunhua Shen, and Anton Van Den~Hengel.
\newblock Wider or deeper: Revisiting the resnet model for visual recognition.
\newblock {\em Pattern Recognition}, 2019.

\bibitem{xie2017adversarial}
Cihang Xie, Jianyu Wang, Zhishuai Zhang, Yuyin Zhou, Lingxi Xie, and Alan
  Yuille.
\newblock Adversarial examples for semantic segmentation and object detection.
\newblock In {\em ICCV}, 2017.

\bibitem{zhang2020causal}
Dong Zhang, Hanwang Zhang, Jinhui Tang, Xiansheng Hua, and Qianru Sun.
\newblock Causal intervention for weakly-supervised semantic segmentation.
\newblock In {\em NeurIPS}, 2020.

\bibitem{zhangsplitting}
Tianyi Zhang, Guosheng Lin, Weide Liu, Jianfei Cai, and Alex Kot.
\newblock Splitting vs. merging: Mining object regions with discrepancy and
  intersection loss for weakly supervised semantic segmentation.
\newblock In {\em ECCV}, 2020.

\bibitem{zhang2018adversarial}
Xiaolin Zhang, Yunchao Wei, Jiashi Feng, Yi Yang, and Thomas~S Huang.
\newblock Adversarial complementary learning for weakly supervised object
  localization.
\newblock In {\em CVPR}, 2018.

\bibitem{zhou2016learning}
Bolei Zhou, Aditya Khosla, Agata Lapedriza, Aude Oliva, and Antonio Torralba.
\newblock Learning deep features for discriminative localization.
\newblock In {\em CVPR}, 2016.

\end{thebibliography}
